\documentclass{article}

\usepackage[nonatbib, main, final]{neurips_2026}

\usepackage[utf8]{inputenc} 
\usepackage[T1]{fontenc}    
\usepackage{url}            
\usepackage{booktabs}       
\usepackage{amsfonts}       
\usepackage{nicefrac}       
\usepackage{microtype}      
\usepackage{xcolor}         

\usepackage{xspace}
\usepackage{tabularray}
\usepackage{tabularx}
\usepackage{multirow}
\usepackage{graphicx}
\usepackage{makecell}
\usepackage{floatrow}
\usepackage{verbatim}
\usepackage{wrapfig}
\usepackage{subcaption}
\usepackage[colorlinks, urlcolor=myblue, linkcolor=myblue, citecolor=gray]{hyperref}

\usepackage{bbm}
\usepackage[most]{tcolorbox}
\usepackage{multicol}
\usepackage{enumitem}
\usepackage{xurl}
\usepackage{tabularray}
\usepackage[htt]{hyphenat}
\usepackage[table]{xcolor}
\usepackage{fontawesome5}
\usepackage[numbers]{natbib}
\usepackage{adjustbox}

\usepackage{tocloft}
\usepackage{titletoc}
\usepackage{etoolbox}
\usepackage{titlesec}

\usepackage{tocbibind}

\UseTblrLibrary{booktabs}           
\DefTblrTemplate{firsthead,middlehead,lasthead}{default}{
}
\DefTblrTemplate{firstfoot}{default}{
  \UseTblrTemplate{contfoot}{default}
  \UseTblrTemplate{caption}{default}
}
\DefTblrTemplate{middlefoot}{default}{
  \UseTblrTemplate{contfoot}{default}
  \UseTblrTemplate{capcont}{default}
}
\DefTblrTemplate{lastfoot}{default}{
  \UseTblrTemplate{note}{default}
  \UseTblrTemplate{remark}{default}
  \UseTblrTemplate{capcont}{default}
}

\newfloatcommand{capbtabbox}{table}[][\FBwidth]
\definecolor{mydarkblue}{rgb}{0,0.08,0.45}
\definecolor{mydarkred}{rgb}{0.75,0,0}
\definecolor{myblue}{HTML}{268BD2}
\definecolor{mygreen}{HTML}{658354}
\definecolor{orangeinplot}{HTML}{e29c7a}
\definecolor{purpleinplot}{HTML}{7676a4}
\definecolor{greeninplot}{HTML}{288308}

\newtcolorbox{prompt}{
  width=\linewidth,
  colback=gray!10!white, 
  colframe=gray!60!black,
  coltext=black,
  left=0.75mm,
  right=0.75mm,
  top=0.7mm,
  bottom=0.4mm,
  enhanced jigsaw,  
  enforce breakable,
}

\newcommand{\papertitle}{Tracing Agentic Failure from the Flow of Success}
\title{\papertitle}

\newcommand{\approach}{\textsc{Oat}\xspace}

\newcommand{\eg}{{\it e.g.}\xspace}
\newcommand{\ie}{{\it i.e.}\xspace}

\ifx\definition\undefined
\newtheorem{definition}{Definition}[section]
\fi

\ifx\theorem\undefined
\newtheorem{theorem}{Theorem}[section]
\fi

\ifx\lemma\undefined
\newtheorem{lemma}{Lemma}[section]
\fi

\definecolor{mypurple}{HTML}{4C3C83}
\definecolor{myyellow}{HTML}{a98467}

\definecolor{goodcolor}{HTML}{2A9D8F}
\definecolor{badcolor}{HTML}{E76F51}

\usepackage{amsmath,amsfonts,bm}

\def\eqref#1{equation~\ref{#1}}

\def\1{\bm{1}}

\DeclareMathAlphabet{\mathsfit}{\encodingdefault}{\sfdefault}{m}{sl}
\SetMathAlphabet{\mathsfit}{bold}{\encodingdefault}{\sfdefault}{bx}{n}

\author{%
  Samuel Yeh\textsuperscript{1}\quad Yiwen Zhu\textsuperscript{2}\quad Shaleen Deep\textsuperscript{2}\quad Sharon Li\textsuperscript{1}\\
  \textsuperscript{1}Department of Computer Science, University of Wisconsin-Madison\quad \\\textsuperscript{2}Microsoft Research\\
\texttt{\{samuelyeh, sharonli\}@cs.wisc.edu}
}

\begin{document}

\maketitle

\begin{abstract}
  Failure attribution for LLM-based agentic systems, \ie, identifying which steps in a failure trajectory caused the task to fail, is critical for debugging and improving these systems. Existing approaches either rely on prompting-based pipelines, which are computationally expensive, or require post-training on failure trajectories with step-level error annotations, which are costly to collect and difficult to scale. We argue that a practical failure attribution model should be lightweight and trainable without step-level supervision on failure data. To this end, we address {{unsupervised failure attribution}}, \ie, training exclusively on successful trajectories and identifying error steps at inference time given a failure trajectory. We propose \approach, which casts this problem as one-class learning with neural controlled differential equations, modeling the dynamical pattern of successful trajectories in latent space. At inference time, each step in a failure trajectory is assigned an anomaly score based on its deviation from the dynamics learned on successful trajectories, which is then used to form a set of error steps. With training on only 100 successful trajectories, experiments show that \approach is \textbf{200--5000$\times$} faster than prompting-based baselines, and, at the same time, consistently outperforms them in both in-domain and out-of-distribution datasets with \textbf{+20\%} and \textbf{+7\%} F1 scores, respectively, demonstrating that \approach is a promising and efficient direction for diagnosing agentic system failures. Code \& Dataset: \href{https://github.com/deeplearning-wisc/OAT}{\faLink}
\end{abstract}

\section{Introduction}

When an LLM-based agentic system fails on a complex, long-horizon task, which of its dozens or hundreds of steps went wrong? Answering this question is extremely challenging: these systems solve tasks by orchestrating multiple specialized agents that interact with tools and external environments~\citep{yao2023react, Wang_2024, 10.24963/ijcai.2024/890}, and a trajectory may span hundreds of actions across agents. Further, the root cause of a failure can be obscured by subsequent steps that partially compensate for earlier mistakes. Without automated tools, diagnosing a single failure trajectory can require a human expert to spend hours manually tracing through the entire trajectory, which is not only prohibitive at scale but also cognitively demanding and prone to inconsistency. As agentic systems are deployed in increasingly high-stakes settings, such as coding~\citep{jimenez2024swebench, wang2025openhands, liu2025largelanguagemodelbasedagents} and scientific research~\citep{lu2026towards, schmidgall-etal-2025-agent, ren2026scientificintelligencesurveyllmbased}, this diagnostic bottleneck becomes a critical barrier to reliable deployment, debugging, and auditing of agentic systems.

To address this challenge, \citet{zhang2025which} introduced the problem of \emph{failure attribution}, identifying the step(s) responsible for the failure given a failure trajectory, and demonstrated that even state-of-the-art reasoning models achieve below 15\% accuracy on this task. Subsequent work developed increasingly sophisticated approaches, which focused on developing sophisticated prompting pipelines~\citep{banerjee2025didwronghierarchicallook, wang2026flatlogscausalgraphs, in2026rethinkingfailureattributionmultiagent} or post-training an LLM with reinforcement learning on failure trajectories with step-level error annotations~\citep{zhang2026agentracer, zhang2025graphtracergraphguidedfailuretracing}. While these methods have shown promise on benchmarks, they suffer from two fundamental limitations. First, annotating step-level error labels on failure trajectories is costly and inherently ambiguous, making it difficult to scale up for real-world use. 
Second, prompting-based approaches require running frontier LLMs at inference time, incurring substantial token costs and latency that preclude deployment in practice.

\begin{figure}
    \centering
    \includegraphics[width=\linewidth]{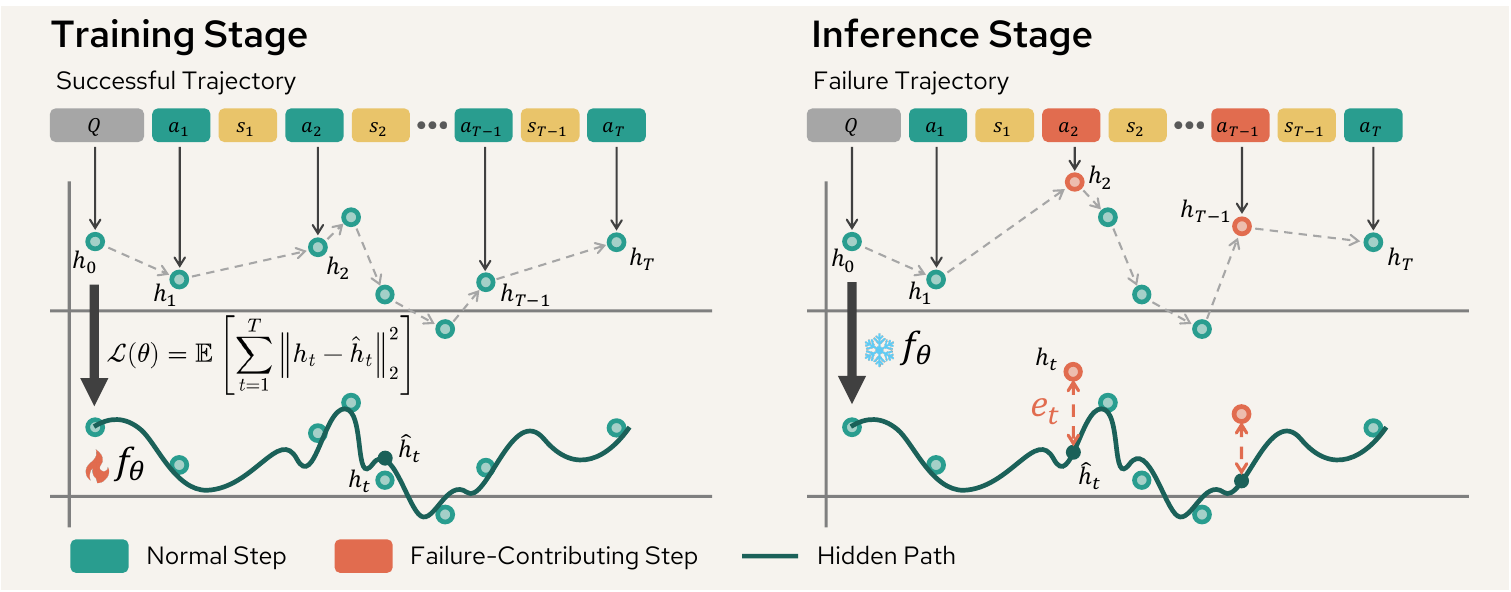}
    \caption{\textbf{Overview of \approach.} \approach learns to model the hidden path and to reconstruct the hidden representations of successful trajectories through Neural Control Differential Equations. At inference time, an expected successful path of a failure trajectory is predicted by \approach, and failure-contributing steps are identified through anomaly scores calculated by the distance between the actual representation and the expected successful path.}
    \label{fig:overview}
    \vspace{-0.2cm}
\end{figure}

 We argue that a practical failure attribution model should be lightweight, deployable without frontier LLM inference, and trainable \emph{without step-level supervision} on failure data.  To achieve this goal, we address a practical problem, \emph{\textbf{unsupervised failure attribution}}: training a model exclusively on successful trajectories and, at inference time, identifying error steps given a failure trajectory. This formulation eliminates the annotation bottleneck, since successful trajectories are easy to collect from any agentic system as a byproduct of normal operation, requiring no step-level labeling effort.

To tackle the problem, we propose \approach 
that casts it as a \textbf{O}ne-class \textbf{A}gent \textbf{T}racing problem in latent representation space. Our key idea is to train a model to characterize the dynamical pattern in successful trajectories, and at inference time, assign each step in a failure trajectory an anomaly score based on its deviation from the learned normal flow. Specifically, we represent each step of a trajectory as a latent vector extracted from an LLM, and model the sequence of step representations as a continuous latent path using Neural Controlled Differential Equations (Neural CDEs)~\citep{10.5555/3495724.3496286}, a principled way to model irregularly structured trajectories. Drawing the connection between Neural CDEs and agentic settings is both new and non-trivial. Latent structure of agentic trajectories reflects goal-directed behavior, and detecting failure requires sensitivity to localized step-level deviations. We establish this connection through two key designs: a continuous latent path over LLM-extracted step representations that captures the temporal dynamics of agent behavior, and a novel gated control path that adaptively suppresses out-of-distribution control signals, improving robustness when the model is deployed on trajectories from unseen domains. 

We train \approach on successful trajectories sampled from {MCP-Atlas}~\citep{bandi2026mcpatlaslargescalebenchmarktooluse}, and evaluate on failure trajectories from both MCP-Atlas as well as {Who\&When}~\citep{zhang2025which}. Despite being trained on only 100 successful trajectories with no failure data and no step-annotation whatsoever, \approach (parameterized by a lightweight 3-layer MLP) consistently outperforms prompting frontier LLMs, including GPT-5. Furthermore, \approach requires zero token cost at inference time and runs \textbf{200--5000$\times$}
 faster than prompting-based approaches, making real-time failure attribution practical for the first time. Our main contributions are as follows:
\begin{enumerate}[leftmargin=15pt, nosep]
    \item We formulate {unsupervised failure attribution} for agentic systems, a new problem setting that eliminates the need for step-level annotation of failure trajectories by training exclusively on successful trajectories.
    \item We propose \approach, a continuous-time one-class learning approach based on neural CDEs with a novel gated control path, modeling the dynamics of successful trajectories in latent space and detecting deviations at inference time.
    \item We conduct extensive experiments and demonstrate that our approach is an effective, practical, and computationally efficient direction for diagnosing agentic system failures.
\end{enumerate}

\section{Related Work}

\vspace{-0.2cm}
\paragraph{Failure attribution for LLM agents.}

The problem of diagnosing failures in LLM-based multi-agent systems has attracted growing attention. Early work focused on characterizing failure modes through taxonomy construction. \citet{cemri2025why} developed the Multi-Agent System Failure Taxonomy, classifying failure modes into 14 categories across system design, inter-agent misalignment, and task verification. \citet{deshpande2025trailtracereasoningagentic} introduced a fine-grained error taxonomy spanning reasoning, planning, and execution failures. 
\citet{kong2026aegis} proposed an RL-based method to identify error-contributing agents and error modes in multi-agent systems.
Beyond failure categorization, recent works shifted focus on localizing error-contributing steps and/or actions~\citep{zhang2025which, in2026rethinkingfailureattributionmultiagent, qian2026actionunveilinginternaldrivers}. In particular, 
\citet{zhang2025which} first proposed to localize the error-contributing step within failure trajectories. They introduced the {Who\&When} benchmark and found that even state-of-the-art reasoning models achieve below 15\% accuracy on localizing the error step. Building on that, many works developed prompting pipelines or learning algorithms to identify error steps~\citep{banerjee2025didwronghierarchicallook, wang2026flatlogscausalgraphs, in2026rethinkingfailureattributionmultiagent, zhang2026agentracer, zhang2025graphtracergraphguidedfailuretracing}. For example, \citet{zhang2026agentracer} post-trained an LLM with GRPO~\citep{shao2024deepseekmathpushinglimitsmathematical} on synthetic failure trajectories. These approaches, however, \emph{require extensive computational resources or large-scale step-level error annotations}, limiting their practicality on real-world deployment. 
There are also intervention-based methods such as \citep{ma2026dover}, which directly intervene in the target trajectory at certain steps and validate whether such interventions recover the failed task. Such methods, however, require more computation to generate additional rollouts, as well as an environment that can be re-simulated at every step, which are not practical in most scenarios. 
The idea of learning from successful trajectories to identify failures has been explored in the robotics community~\citep{xu2025can, romer2026failure, zhou2026rcnfrobotconditionednormalizingflow}, where one-class models are trained on normal executions and anomalous trajectories are flagged at inference time. However, these works focus on trajectory-level detection, \ie, distinguishing successful from failed executions as a whole, rather than localizing the specific steps that caused the failure. Our work bridges this gap by bringing the unsupervised, one-class learning paradigm into the LLM agentic setting and extending it to {step-level} failure attribution, a strictly harder and more informative task that has not been attempted in the agentic AI community.

\vspace{-0.2cm}
\paragraph{Anomaly detection.}

Anomaly detection is a well-studied problem in machine learning, aiming to identify observations that deviate from learned normal behavior~\citep{10.1016/j.cose.2008.08.003, chalapathy2019deeplearninganomalydetection, 10.1145/3691338}. One-class learning methods train exclusively on normal data and identify anomalies as out-of-distribution samples at test time~\citep{NIPS1999_8725fb77, 10.1023/B:MACH.0000008084.60811.49, Hayashi_2026, pmlr-v80-ruff18a, 10.1145/2689746.2689747, xu2022anomaly, park2018multimodal, 10.1145/3292500.3330672}. 
However, standard anomaly detection focuses on identifying whether a given sample is anomalous at the sample-level. 
In agent settings, this amounts to classifying whether a trajectory succeeded or failed, a coarse signal that reveals nothing about where a failure occurred. In this paper, we introduce a novel formulation of one-class learning for step-level failure attribution in agent trajectories. Rather than classifying trajectories at inference time, we derive anomaly score for each step in a failure trajectory and use it to localize failure-contributing steps.

\section{Problem Statement}\label{sec:problem_statement}

An agentic system consists of multiple LLM-based agents, each equipped with tool-calling and inter-agent communication capabilities, collectively designed to solve a complex, long-horizon task. At each time step $t$, the system observes a state $s_t$ from an environment and chooses an agent to execute an action $a_t$. 
Given a query $Q$, a \emph{trajectory} is denoted by
\begin{align}
    \tau = (Q, a_1, s_1, a_2, \dots, s_{T_\tau-1}, a_{T_\tau}),
\end{align}
where $T_\tau$ is the terminal step, and we write $\tau_{<t} = (Q, a_1, s_1, \dots, s_{t-1})$ for the prefix preceding step $t$. 
An evaluation function $Z(\tau) \in \{0, 1\}$ determines whether the trajectory successfully completes the task ($Z(\tau) = 1$) or not ($Z(\tau) = 0$) based on its final outcome, following prior work~\citep{zhang2025which}. $Z$ can be computed automatically by the environment for verifiable tasks, or assessed by humans for subjective ones. 

Given a failure trajectory $\tau'$ with $Z(\tau') = 0$, \emph{failure attribution} is the task of identifying which step(s) caused the failure, formalized as predicting a set of error steps $y(\tau') \subseteq \{1, \dots, T_{\tau'}\}$. Existing works~\citep{zhang2026agentracer} have introduced \emph{supervised failure attribution}, which requires training on a set of step-annotated failure trajectories. However, existing supervised approaches have two limitations. First, collecting step-level annotations is both costly and inherently ambiguous due to the reliance on an oracle rectification function that is inaccessible during practical annotation~\citep{ma2026dover, in2026rethinkingfailureattributionmultiagent}. Second, they focus on single-step interventions and therefore cannot capture compound errors, where multiple steps jointly contribute to failure. We address these limitations with two components: an operational notion of which steps constitute errors (Definition~\ref{def:fcs}), and a learning setting that requires no step-level annotations (Definition~\ref{def:ufa}).

\vspace{-0.2cm}
\begin{definition}[Failure-Contributing Steps]
\label{def:fcs}
Given a failure trajectory $\tau'$ with $Z(\tau') = 0$, a step $t$ is \emph{failure-contributing} if (i) $a_t$ is erroneous given $Q$ and $\tau'_{<t}$, \ie, it contains a hallucination, makes an erroneous tool call, or fails to handle a tool error in $\tau'_{<t}$; and (ii) the error is not corrected later and contributes to the failure. We denote the set of such steps by $y(\tau') \subseteq \{1, \dots, T_{\tau'}\}$.
\end{definition}

Unlike the counterfactual criterion of~\citet{zhang2025which}, Definition~\ref{def:fcs} does not require that correcting step $t$ alone would recover the task. This choice reflects two observations. First, failures in complex agentic tasks often arise from multiple compounding errors. Second, the counterfactual criterion is hard to verify in practice, since annotators can neither obtain an oracle replacement action nor regenerate the trajectory from the corrected step~\citep{ma2026dover}. We further describe how we adopt this definition during dataset curation and label annotation in Appendix~\ref{ap:dataset} and \ref{ap:annotation}.

\vspace{-0.2cm}
\begin{definition}[Unsupervised Failure Attribution]
\label{def:ufa}
    Let $\mathcal{D}_{\text{succ}} = \{\tau^{(k)}\}$ be a dataset of successful trajectories with no step-level annotations, \ie, $Z(\tau^{(k)}) = 1$ for all $\tau^{(k)} \in \mathcal{D}_{\text{succ}}$. The goal is to learn a predictor $\psi_\theta$, trained exclusively on $\mathcal{D}_{\text{succ}}$, that at inference time identifies the failure-contributing steps of a failure trajectory $\tau'$:
    \begin{align}
        \psi_\theta(\tau') \approx y(\tau').
    \end{align}
\end{definition}

Note that Definition~\ref{def:ufa} differs from supervised failure attribution~\citep{zhang2026agentracer} in its training signal. Supervised methods learn from annotated pairs $(\tau', y(\tau'))$. In our setting, the predictor observes neither failure trajectories nor step labels during training. Accordingly, the learning objective in the next section is defined solely over $\mathcal{D}_{\text{succ}}$, and $y(\tau')$ is used only for evaluation.

\section{Methodology}\label{sec:method}

In this work, we frame unsupervised failure attribution as a \emph{one-class learning problem} in LLM's representation space, which contains rich information beyond generated text: we train a model to characterize the dynamical pattern in successful trajectories, and at inference time identify steps in failure trajectories that deviate from the normal flow as contributing steps.

Concretely, given a trajectory $\tau = (Q, a_1, s_1, a_2, \dots, s_{T_\tau-1}, a_{T_\tau})$ and the LLM agent $M_t$ active at step $t$, we define the representation of step $t$ as the aggregated token representations produced by layer $\ell$ of $M_t$ when generating action $a_t$:
\begin{align}
    h_t = M_t^{(\ell)}(a_t \mid Q, a_1, s_1, \dots, a_{t-1}, s_{t-1}) \in \mathbb{R}^{d_h}.
\end{align}
Multiple aggregation strategies are possible within a single step, such as using the last token or averaging across all tokens; we defer this discussion to Appendix~\ref{ap:ablation}.
We also define the query representation as $h_Q=M_1^{(\ell)}(Q)$, and denote the full sequence of representations as $H(\tau) = (h_1, \dots, h_{T_\tau}) \in \mathbb{R}^{T_\tau \times d_h}$. 
Next, we describe how we model the dynamics $H(\tau)$ in Section~\ref{sec:cde} and detect the failure steps in Section~\ref{sec:score}.

\subsection{Continuous Trajectory Modeling}\label{sec:cde}

Agent behavior is inherently a continuous dynamic process: each action shapes the subsequent state in a temporally coherent way. Given a successful trajectory $\tau$, we treat the query representation $h_{Q}$ together with the step representations $H(\tau) = (h_1, \dots, h_{T_\tau})$ as irregularly sampled observations of an underlying continuous latent path, where the system evolves from the initial query state $h_{Q}$ toward the terminal goal state $h_T$. Formally, we normalize all trajectories to the unit interval $[0, 1]$ via a monotonic mapping $u: \{0, 1, \dots, T_\tau\} \to [0, 1]$ with $u(t) = t/T_\tau$, and re-index the representations as $h_{u_0}, h_{u_1}, \dots, h_{u_{T_\tau}}$ with $u_t = u(t)$ and $h_{u_0} = h_{Q}$. Note that the index $u_t$ is now continuous.

\paragraph{Trajectory dynamics modeling.} Under this formulation, the intermediate representations $(h_{u_1}, \dots, h_{u_{T_\tau}})$ are observations sampled along the latent trajectory, and modeling their continuous evolution is the key technical challenge. Discrete-time models such as RNNs and Transformers treat these observations as isolated events and do not capture the continuous dynamics between them. Neural Differential Equations (NDEs)~\citep{10.24963/ijcai.2025/1179} offer a principled alternative for continuous-time modeling with neural networks, and are particularly well-suited to agentic trajectories, which are variable-length and irregularly structured.
A natural starting point within the NDE family is Neural ODEs~\citep{NEURIPS2018_69386f6b, 10.5555/3454287.3454950, NEURIPS2019_99a40143, 10.5555/3454287.3454765}, which model the evolution of a latent state  from an initial condition:
\begin{align}
    \hat{h}(u)=g_1(z(u); \theta_{g_1}),\quad z(u) = z(0) + \int_0^u f\!\left(v, z(v);\, \theta_f\right) \mathrm{d}v, \quad z(0)=g_2(h_Q; \theta_{g_2}),\label{eq:ode}
\end{align}
where $g_1:\mathbb{R}^{d_z}\to\mathbb{R}^{d_h}$ and $g_2:\mathbb{R}^{d_h}\to\mathbb{R}^{d_z}$ are linear maps, $h_Q$ is an initial state, $\hat{h}:[0,1]\to\mathbb{R}^{d_h}$ and $z:[0,1]\to\mathbb{R}^{d_z}$ are continuous functions of representations, and $f$ is a lightweight neural network (e.g., MLP) approximating $\mathrm{d}z(v)/\mathrm{d}v$. We can train $g_1, g_2,$ and $f$ so that the reconstructed representation $\hat{h}(u_t)$ matches the representation of successful trajectories at each step. Formally, the training objective is:
\begin{align}
    \mathcal{L}(\theta) = \mathbb{E}_{\tau\sim\mathcal{D}_{\text{succ}}}\left[\sum_{t=1}^{T_\tau} \left\| h_{t} - 
    \hat{h}\!\left(u_t\right) \right\|_2^2\right].\label{eq:loss}
\end{align}

One limitation of Neural ODEs is that the latent trajectory is \emph{driven solely by the initial condition $h(0)$ and does not take subsequent observations into account}. As a result, the learned dynamics define a deterministic flow that maps each initial state to a single trajectory. This is particularly limiting in agentic systems, where multiple distinct action sequences may all lead to successful outcomes. To address this issue, we consider its variant, Neural Controlled Differential Equations (Neural CDEs)~\citep{10.5555/3495724.3496286, morrill2021neuralcontrolleddifferentialequations}, which resolve this limitation by continuously conditioning the latent dynamics on the observed trajectory. A continuous control path $X:[0,1]\to\mathbb{R}^{d_h+1}$ is constructed by interpolating the discrete step representations via  cubic spline, with $X(u_t)=(h_t, u_t)$, whose derivative $\mathrm{d}X(v)/\mathrm{d}v$ drives the latent dynamics at every integration step:
\begin{align}
     z(u)= z(0) + \int_0^u f\!\left(v, z(v);\, \theta_f\right) \mathrm{d}X(v),
\end{align}
where the integral is interpreted as a Riemann-Stieltjes integral, and can be rewritten as 
\begin{align}
    \int_0^u f\!\left(v, z(v);\, \theta_f\right) \mathrm{d}X(v) = 
    \int_0^u f\!\left(v, z(v);\, \theta_f\right) \frac{\mathrm{d}X(v)}{\mathrm{d}v}\, 
    \mathrm{d}v.\label{eq:ncde_2}
\end{align}
Note that $X$ is not learnable. Eq. (\ref{eq:ncde_2}) indicates that, rather than relying only on a fixed initial condition, $f$ is continuously modulated by $\mathrm{d}X(v)/\mathrm{d}v$ at every integration step, which encodes the direction and rate of change of the observations. As a result, the latent state $\hat{h}(u)$ is sensitive to the local behavior of the trajectory at each step, making Neural CDEs well-suited to detecting localized deviations in failure trajectories. We adopt the same $g_1$ and $g_2$ to project $z(u)$ and $h_Q$, respectively, and use the same loss to train them as defined in Eq. (\ref{eq:loss}).

\paragraph{Gated control path.} 

One practical challenge of deploying Neural CDEs in a real-world setting is that the control path $X(u)$ may carry spurious signals when the input trajectory is out-of-distribution. Since the model is trained exclusively on successful trajectories from certain domains, the magnitudes of the spline derivatives $\mathrm{d}X(u)/\mathrm{d}u$ at test time may deviate significantly from those seen during training, causing the control path to inject misleading dynamics into the latent state rather than informative guidance. To address this, we propose a \emph{gated control path} that adaptively regulates the influence of the control signal. Specifically, we introduce a gating function $q: \mathbb{R}^{d_h+1} \to [0,1]^{d_h+1}$ parameterized by a neural network, and replace the standard control path derivative with a gated variant:
\begin{align}
    \frac{\mathrm{d}\tilde{X}(u)}{\mathrm{d}u} = q\!\left(\frac{\mathrm{d}X(u)}{\mathrm{d}u};\theta_q\right) \odot 
    \frac{\mathrm{d}X(u)}{\mathrm{d}u},\label{eq:gate}
\end{align}
where $\odot$ denotes element-wise multiplication. When the trajectory is in-distribution, $q$ can pass the control signal through at full strength; when the trajectory is out-of-distribution and the spline derivatives are atypically large or directionally unfamiliar, $q$ suppresses them, reducing the risk of the control path destabilizing the latent dynamics. The modified CDE then evolves as:
\begin{align}
    z(u) = z(0) + \int_0^u f\!\left(v, z(v);\, \theta_f\right) 
    \frac{\mathrm{d}\tilde{X}(v)}{\mathrm{d}v}\, dv.
\end{align}
This design preserves the full expressiveness of Neural CDEs in the in-domain setting while improving robustness to distributional shift, as confirmed by our ablation study in Section~\ref{sec:ablation}.

\subsection{Detecting Failure-Contributing Steps}
\label{sec:score}

\paragraph{Step-level anomaly score.} At inference time, given a failure trajectory $\tau'$ with step representations $H(\tau') = (h_Q', h_1', \dots, h_{T_{\tau'}}')$, we pass it into $f$ and produce the reconstructed states $\hat{h}(u_t)$ for each step $t$. The anomaly score for each step $t$ is then the reconstruction error:
\begin{align}
    e_t = \left\| h_{t}' - \hat{h}\!\left(u_t\right) \right\|_2^2.\label{eq:anomaly_score}
\end{align}
Since $f$ is trained exclusively on successful trajectories, it learns to predict the expected dynamics of a well-functioning system. Steps in a failure trajectory where the actual latent state deviates substantially from the predicted dynamics will yield high anomaly scores.

Given the per-step anomaly scores $\{e_t\}_{t=1}^{T_{\tau'}}$ produced by the model, the final step is to identify which steps constitute the set of failure-contributing steps $y(\tau')$. We consider two detection strategies.

\paragraph{Top-$k$ detection.}
The simplest approach is to select the $k$ steps with the highest anomaly scores:
\begin{align}
    \hat{y}(\tau') = \left\{ t \;\middle|\; e_t \in \text{top-}k\!\left(
    \{e_1, \dots, e_{T_{\tau'}}\}\right) \right\},
\end{align}
where $k$ is a fixed hyperparameter. This strategy is straightforward but requires prior knowledge of how many failure-contributing steps a typical failure trajectory contains, which may vary across tasks and systems.

\paragraph{Conformal prediction (CP) detection.}
To obtain a more principled and adaptive threshold, we adopt conformal prediction~\citep{10.1561/2200000101}, which provides distribution-free coverage guarantees without assumptions on the underlying data distribution. Specifically, we construct a calibration set $\mathcal{D}_{\text{cal}} \subset \mathcal{D}_{\text{succ}}$ of held-out successful trajectories. For each calibration trajectory $\tau \in \mathcal{D}_{\text{cal}}$, we compute the per-step anomaly scores and treat them as nonconformity scores measuring how well each step conforms to the learned distribution. The threshold $\delta$ is then set as the $(1-\alpha)$-quantile of the calibration scores:
\begin{align}
    \delta = \operatorname{Quantile}\!\left(1 - \alpha,\; 
    \left\{ e_t^{(\tau)} \right\}_{\tau \in \mathcal{D}_{\text{cal}},\, 
    t \in \{1,\dots,T_\tau\}} \right),
\end{align}
where $\alpha \in (0, 1)$ is a user-specified miscoverage rate. At inference time, steps in the failure trajectory whose anomaly scores exceed $\delta$ are flagged as failure-contributing steps:
\begin{align}
    \hat{y}(\tau') = \left\{ t \mid e_t > \delta \right\}.
\end{align}
This strategy adaptively determines the number of failure-contributing steps based on the learned distribution of normal behavior, and provides a formal guarantee that the false positive rate on in-distribution steps is controlled at level $\alpha$.

\section{Experiments}\label{sec:experiment}

\subsection{Experimental Setup.}\label{sec:setup}

\paragraph{Datasets.}

We construct the training and in-domain evaluation set by sampling trajectories from \textbf{MCP-Atlas}~\citep{bandi2026mcpatlaslargescalebenchmarktooluse}. MCP-Atlas is a benchmark for tool-calling agents with claim-level evaluation. We sample trajectories with Qwen3.5-27B~\citep{qwen3.5} and consider trajectories with no claim-level error as successful trajectories. We manually annotate the set of contribution steps for failure trajectories. In total, we obtain 103 successful and 88 failure trajectories. We use \textbf{Who\&When}~\citep{zhang2025which} for out-of-distribution (OOD) evaluation, which contains 184 step-annotated failure trajectories sampled with GPT-4o~\citep{openai2024gpt4ocard} on tasks 
drawn from GAIA~\citep{mialon2024gaia} and AssistantBench~\citep{yoran-etal-2024-assistantbench}.
We provide the details of these two datasets and annotation process in Appendix~\ref{ap:dataset} and \ref{ap:annotation}, respectively. 
Note that the label semantics are different in the two datasets. For MCP-Atlas, the label is a set of failure-contributing steps. Whereas for Who\&When, the label is the index of the earliest decisive error. To account for this difference, we evaluate \approach with set- and rank-based metrics as described next.

\paragraph{Evaluation metrics.}

We consider the following set-based metrics: (1) \textbf{Precision}: $\sum_{t\in\hat{y}(\tau)}\mathbb{I}[t\in y(\tau)]/|\hat{y}(\tau)|$, (2) \textbf{Recall}: $\sum_{t\in y(\tau)}\mathbb{I}[t\in \hat{y}(\tau)]/|y(\tau)|$, (3) \textbf{F1 score}: harmonic mean of precision and recall, and (4) \textbf{Hit rate}: $\mathbb{I}[|\hat{y}(\tau)\cap y(\tau)|>0]$. 
For MCP-Atlas, these metrics measure region overlaps with any labeled step, and for Who\&when, they measure whether the region contains the first decisive step.
We also consider decoder-agnostic metrics, including (5) \textbf{AUROC} and (6) \textbf{AUPRC}, both evaluate the rank of the anomaly scores 
and are less sensitive to the difference between the label of a single step and a set.
All the metrics are reported as the average across all trajectories.

\paragraph{Baselines.}

We compare \approach with prompt-based approaches, which identify failure-contributing steps by prompting an LLM. We evaluate two different LLMs, including GPT-4o~\citep{openai2024gpt4ocard} and GPT-5~\citep{singh2025openaigpt5card}. The prompt is modified from the all-at-once prompt in \citet{zhang2025which} to support predicting a set of failure-contributing steps (See Appendix~\ref{ap:prompt} for detailed prompt design). We also compare it with the random and first-step baselines. For these two baselines, we select only one step for each trajectory. 
On MCP-Atlas, we further compare \approach against two classical anomaly detection methods, OC-SVM~\citep{3009657.3009740} and Isolation Forest~\citep{4781136}. For these two baselines, we score each step independently.

\begin{table*}[t]
    \centering
    \small 
    \resizebox{\textwidth}{!}{
    \begin{tabular}{l cccccc}
        \toprule
         Approach & Precision & Recall & F1  & Hit & AUROC & AUPRC\\
        \midrule
        Random-Step & $0.284$ & $0.246$ & $0.255$ & $0.284$ & $0.529$ & $0.226$\\
        First-Step & $0.136$ & $0.108$ & $0.116$ & $0.136$ & $0.469$ & $0.206$\\
        \midrule
        OC-SVM & $0.311$ & $0.695$ &  $0.411$ & $0.772$  & $0.569$ & $0.282$\\
        Isolation Forest & $0.314$ & $0.691$ & $0.411$ & $\mathbf{0.784}$ & $0.589$ & $0.295$\\
        \midrule
        GPT-4o & $0.233$ 
& $0.210$ 
& $0.212$ 
& $0.250$ 
& $0.509$ 
& $0.217$\\
        GPT-5 & $0.217$ 
& $0.183$ 
& $0.181$ 
& $0.250$ 
& $0.515$ 
& $0.217$\\
        \midrule
        \rowcolor{goodcolor!30}\approach (Top-$k$) 
        & $0.321_{\pm 0.006}$ 
        & $\mathbf{0.706_{\pm 0.015}}$ 
        & $0.420_{\pm 0.008}$ 
        & $0.777_{\pm 0.012}$ 
        & & \\
        
        \rowcolor{goodcolor!30}\approach (CP) 
        & $\mathbf{0.443_{\pm 0.019}}$ 
        & $0.484_{\pm 0.018}$ 
        & $\mathbf{0.435_{\pm 0.013}}$ 
        & $0.566_{\pm 0.024}$   & \multirow{-2}{*}{$\mathbf{0.629_{\pm 0.007}}$} 
        & \multirow{-2}{*}{$\mathbf{0.324_{\pm 0.009}}$} \\
        \bottomrule
    \end{tabular}
    }
    \vspace{-0.2cm}
    \caption{\textbf{\approach outperforms baselines across metrics in an in-domain scenario.} We train and evaluate models on MCP-Atlas. The highest score is marked as \textbf{bold}.
    }
    \vspace{-0.2cm}
    \label{tb:main_result}
\end{table*}

\paragraph{Implementation details.}
We extract step representations using Qwen3.5-27B, 
taking the last-layer hidden states and applying mean-pooling to aggregate token representations within each step. We use Qwen3.5-27B as it is the same model used to simulate trajectories on {MCP-Atlas}. 
For the OOD experiment on Who\&When, where the trajectories are generated by GPT-4o, we again use Qwen3.5-27B as a proxy for obtaining step representations.
To improve generalizability and reduce computational cost, we apply PCA to project the representations to $d_h=64$ dimensions. 
The Neural CDE vector field $f$ is parameterized by a 3-layer MLP with hidden dimension 64, and the control path $X$ is constructed via natural cubic spline interpolation over the sequence of step representations. We adopt Euler's method as an CDE solver. For the two detection strategies, we set $k = 3$ for top-$k$ detection and miscoverage rate $\alpha = 0.2$ for conformal prediction detection. We train Neural CDE on 80\% of successful trajectories and use the rest of 20\% trajectories as validation/calibration set for early stopping and determining threshold of conformal prediction.
Other implementation details and hyperparameters can be found in Appendix~\ref{ap:implementation}.

\subsection{Experimental Results}\label{sec:result}

All experiments based on \approach are run with 5 different random seeds. We report the mean and standard deviation of each metric across the 5 runs.

\paragraph{In-domain evaluation.}

Table~\ref{tb:main_result} presents the in-domain evaluation results on MCP-Atlas. \approach consistently outperforms all baselines across metrics with a huge performance gap (\eg, +\textbf{20}\% F1 score and +10\% AUPRC). Notably, the model is trained on fewer than 100 successful trajectories, yet still surpasses frontier LLMs on failure-contributing steps identification, suggesting that one-class learning on successful trajectories is a promising and label-efficient direction for failure attribution. The results also reveal a clear trade-off between the two detection strategies. Top-$k$ detection achieves higher recall and hit rate but lower precision, indicating a tendency to over-detect failure-contributing steps due to its fixed detection size. Conformal prediction detection, by contrast, achieves higher precision and a better balance between precision and recall, demonstrating the benefit of its adaptive, distribution-calibrated threshold.

\paragraph{OOD evaluation.}

There are three types of mismatches between MCP-Atlas and Who\&When. First, MCP-Atlas focuses on tool-calling in single-agent settings, whereas {Who\&When} evaluates multi-agent collaboration, yielding trajectories with 
fundamentally different structures. Second, MCP-Atlas trajectories are generated by Qwen3.5-27B, while {Who\&When} trajectories are generated by GPT-4o, two models that may exhibit different failure behaviors. Third, Qwen3.5-27B and GPT-4o have different output styles, resulting in different distributions of step representations. In this evaluation, we focus on the first two OOD scenarios.
To mitigate the impact of the third type of mismatch, we apply Correlation Alignment (CORAL)~\citep{sun2016correlationalignmentunsuperviseddomain}, which aligns the second-order statistics of the OOD representations to those of the in-domain representations before they are passed to the Neural CDE. CORAL is fit using only representations from Who\&When trajectories, without accessing task outcomes, failure labels, or step-level annotations.

Table~\ref{tb:ood_result} presents the OOD evaluation results on {Who\&When}. \approach again outperforms prompting-based approaches with +7\% F1 score and +5\% AUPRC, underscoring its generalizability. This is particularly encouraging given the substantial distributional differences between the two benchmarks. 
We believe that it is because \approach models the dynamics rather than specific behaviors. Although MCP-Atlas and Who\&When differ substantially in tasks and underlying LLMs, successful trajectories exhibit common structural patterns, such as coherent reasoning and consistent progress toward the goal. \approach learns these dynamics and identifies deviations from them, making it less dependent on the specific task or model than supervised classifiers trained to recognize particular failure patterns.
These results also highlight a practical advantage of the unsupervised training paradigm: since training requires only successful trajectories, which are easier to collect than annotated failure trajectories, it is straightforward to augment the training set with additional in-distribution data before deploying the model in a new OOD setting, without any additional annotation effort.

\begin{table*}[t]
    \centering
    \small 
    \resizebox{\textwidth}{!}{
    \begin{tabular}{l cccccc}
        \toprule
         Approach & Precision & Recall & F1  & Hit & AUROC & AUPRC\\
        \midrule
        Random-Step & $0.137$ & $0.137$ & $0.137$ & $0.137$ & $0.547$ & $0.061$\\
        First-Step & $0.115$ & $0.115$ & $0.115$ & $0.115$ & $0.535$ & $0.057$\\
        \midrule
        GPT-4o & $0.129$ 
& $0.198$ 
& $0.151$ 
& $0.198$ 
& $0.566$ 
& $0.066$\\
        GPT-5 & $0.111$ 
& $0.275$ 
& $0.152$ 
& $0.275$ 
& $0.584$ 
& $0.078$\\
        \midrule
        \rowcolor{goodcolor!30}\approach (Top-$k$) 
        & $0.150_{\pm 0.002}$ 
        & $\mathbf{0.451_{\pm 0.006}}$ 
        & $\mathbf{0.225_{\pm 0.003}}$ 
        & $\mathbf{0.451_{\pm 0.006}}$ 
        & & \\
        
        \rowcolor{goodcolor!30}\approach (CP) 
        & $\mathbf{0.184_{\pm 0.005}}$ 
        & $0.330_{\pm 0.016}$ 
        & $0.211_{\pm 0.006}$ 
        & $0.330_{\pm 0.016}$   & \multirow{-2}{*}{$\mathbf{0.758_{\pm 0.005}}$} 
        & \multirow{-2}{*}{$\mathbf{0.128_{\pm 0.004}}$}\\
        \bottomrule
    \end{tabular}
    }
    
    \caption{\textbf{\approach achieves a comparable performance or even outperforms baselines across metrics in an OOD scenario.} We train models on MCP-Atlas and evaluate them on Who\&When. The highest score is marked as \textbf{bold}. 
    }
    
    \label{tb:ood_result}
\end{table*}

\begin{wraptable}{r}{0.6\textwidth}
    \centering
    \small
    \resizebox{\textwidth}{!}{
    \begin{tabular}{lcccc}
        \toprule
        & \multicolumn{2}{c}{MCP-Atlas} & \multicolumn{2}{c}{Who\&When}\\
        Approach & \# Tokens $\downarrow$ & Latency (ms) $\downarrow$ & \# Tokens $\downarrow$ & Latency (ms) $\downarrow$ \\
        \midrule
        GPT-4o & $121_{\pm 31}$ & $4241_{\pm 1684}$ & $118_{\pm 33}$ & $4522_{\pm 2351}$\\
        GPT-5 & $3012_{\pm 683}$ & $39626_{\pm 11079}$ & $2694_{\pm 808}$ & $37819_{\pm 14281}$\\
        \midrule
        \rowcolor{goodcolor!30}\approach & $\mathbf{0}$ & $\mathbf{7_{\pm 4}}$ & $\mathbf{0}$ & $\mathbf{16_{\pm22}}$\\
        \bottomrule
    \end{tabular}
    }
    \vspace{5pt}
    \caption{\textbf{\approach costs much less than prompting-based baselines.} We report the number of output tokens and detection latency for each approach. The lowest cost is marked as \textbf{bold}.}
    \label{tb:cost}
\end{wraptable}

\paragraph{Computational efficiency.}

Beyond performance, one additional strength of \approach is its computational cost. Unlike prompting-based approaches that require running frontier LLMs, which can not be run on a local server, the deployment of \approach requires less than 1 GB VRAM, making it easily to deploy. In addition, Table~\ref{tb:cost} reports the number of output tokens and latency for each approach. The result shows that \approach is much efficient than prompting-based approaches in terms of money and time, with 0 token cost and \textbf{200--5000}$\times$ faster,\footnote{
The reported latency of \approach includes every stage after the generator LLM forward pass, except hidden-state extraction, as they are byproducts during generation and does not take extra time to collect. The latency with hidden-state extraction is around 1000ms per trajectory when running with 2 H200 GPUs, which is still 4-35$\times$ faster than prompting baselines.
} allowing a real-time failure attribution.

\begin{figure}
    \centering
    \includegraphics[width=\linewidth]{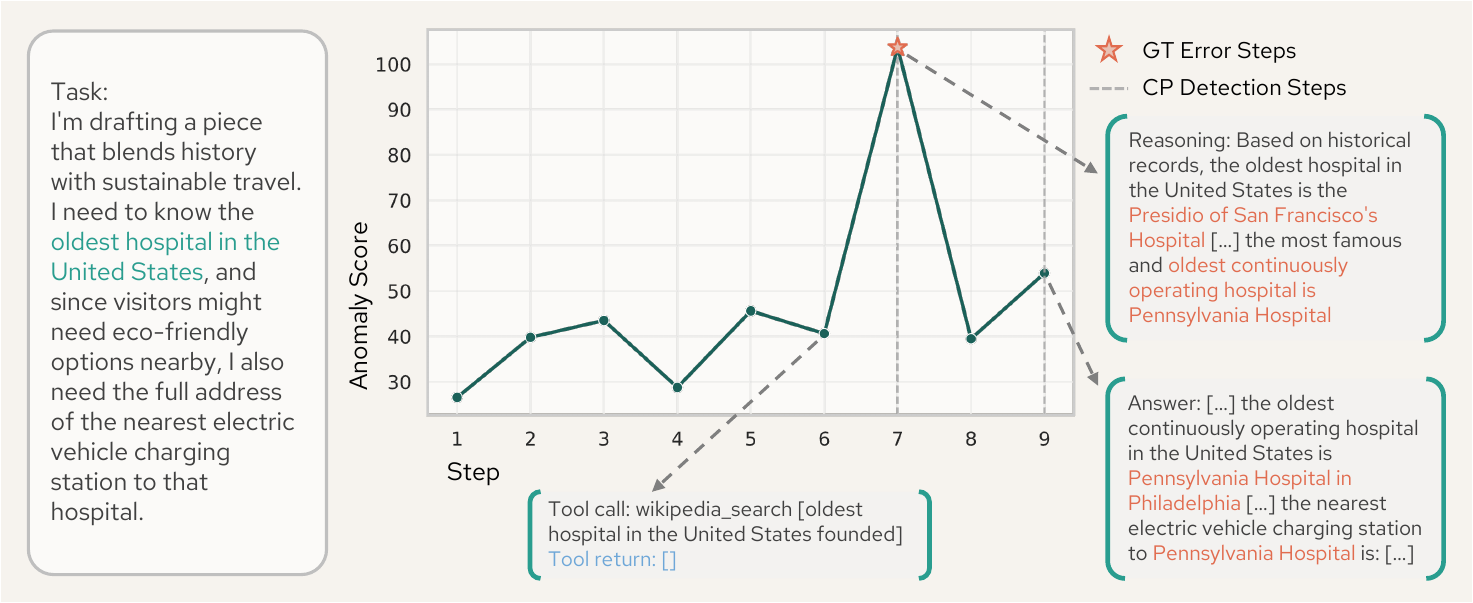}
    \caption{\textbf{Visualization of \approach's output.} \approach correctly identifies the hallucination at step 7 and the propagated error at step 9.}
    \label{fig:success_1}
\end{figure}

\paragraph{Qualitative case studies.} We conduct case studies on MCP-Atlas with 6 randomly selected trajectories.~
Figure~\ref{fig:success_1} shows a case where \approach successfully identified the failure step. Details of additional cases, including failure cases, are shown in Appendix~\ref{ap:case_study}.  Our qualitative analysis reveals several consistent patterns. In successful cases, \approach assigns significantly high anomaly scores to failure-contributing steps, including hallucinations, unfaithful assumptions, and code bugs, while keeping scores for benign steps low. In failure cases, \approach misses the annotated root cause but still surfaces meaningful signals. It assigns progressively increasing scores to suboptimal plans that precede the final failure, and reliably detects steps where the failure becomes explicitly verbalized. Overall, the case studies suggest that \approach learns a reasonable notion of trajectory normality and that more adaptive detection strategies could further improve attribution coverage.

\section{Ablation Studies}\label{sec:ablation}

We conduct ablation studies to analyze the impact of each component of \approach. All the ablations are conducted on MCP-Atlas with conformal prediction detection unless explicit mention.

\begin{wrapfigure}[13]{r}{0.4\textwidth}
\centering
    \includegraphics[width=\textwidth]{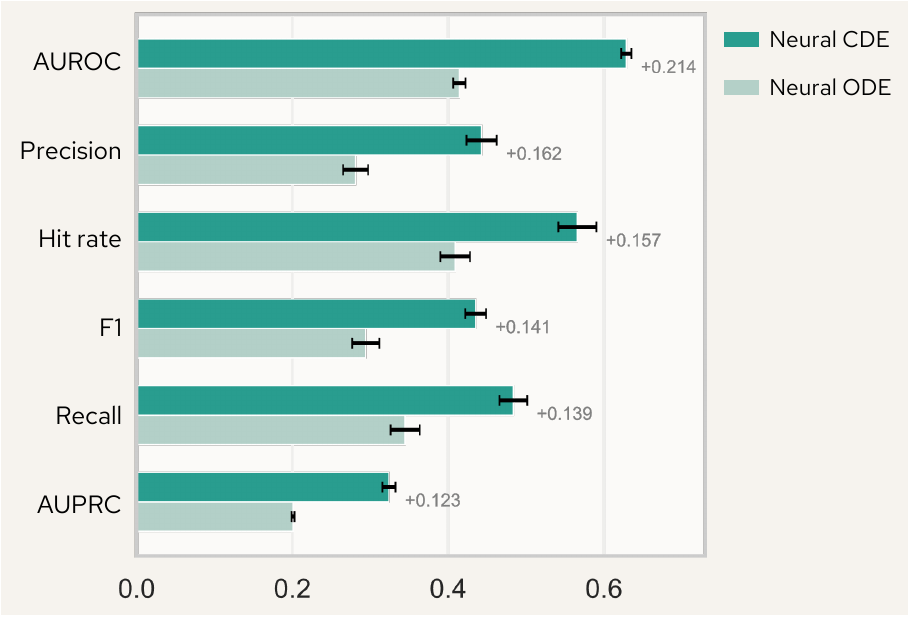}
    \caption{\textbf{Neural CDE consistently outperforms Neural ODE across metrics.}}
    \label{fig:cde_vs_ode}
\end{wrapfigure}
\paragraph{Neural ODE vs. CDE.}

Figure~\ref{fig:cde_vs_ode} shows the performance between Neural ODE and CDE with a non-trivial gap. The result shows that Neural CDE outperforms Neural ODE across all metrics, suggesting that the control path, which continuously injects trajectory observations into the latent dynamics, plays an important role in accurately modeling agent trajectories.

\paragraph{Impact of gated control path.}

\begin{figure}
    \centering
    \includegraphics[width=\linewidth]{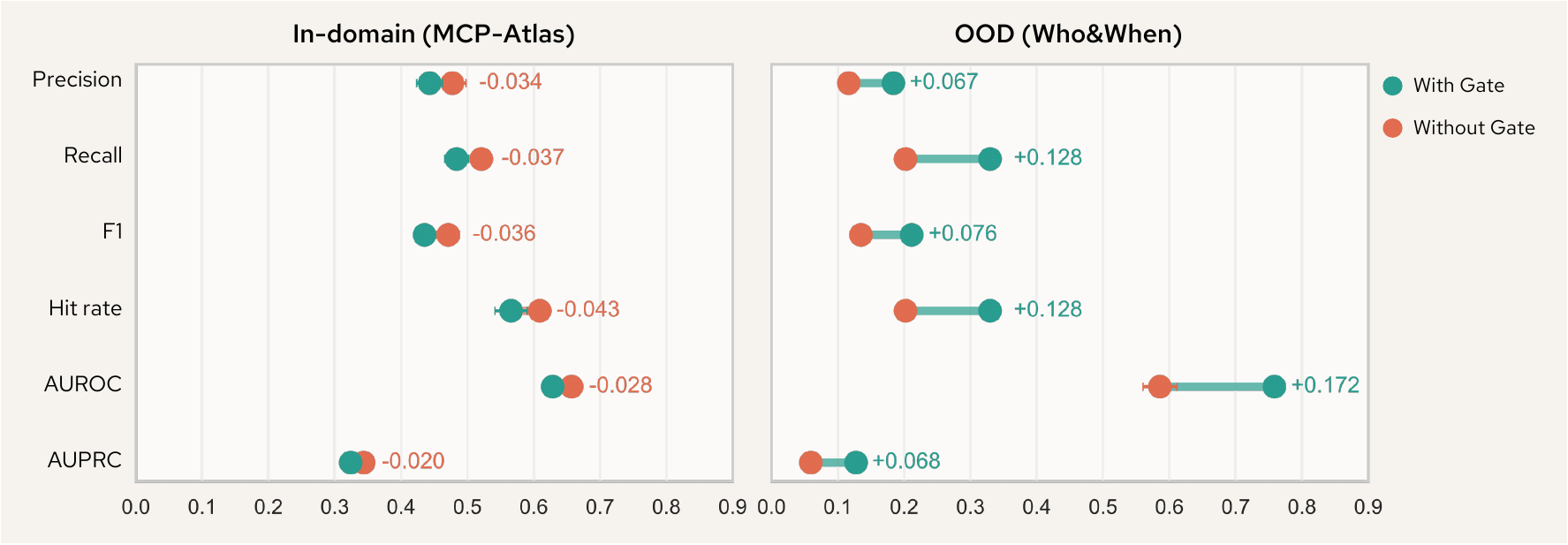}
    \caption{\textbf{Gated control path improves generalizability with a small trade-off on in-domain performance.} We plot the performance of with and without gating function, as well as their gaps, in both in-domain and OOD scenarios. 
    }
    \label{fig:w_vs_wo_gate}
\end{figure}

In Section~\ref{sec:cde}, we replace the original control path in CDE with a gated control path (Eq. (\ref{eq:gate})). We ablate on this design choice to show the impact of the gating function. Figure~\ref{fig:w_vs_wo_gate} shows that gated control path significantly improve the OOD performance by $\mathbf{+0.172}$ AUROC, with a small trade-off ($-0.028$) on in-domain performance. This asymmetric effect suggests that the gating mechanism improves the generalizability of the learned trajectory dynamics, making the model more robust to the distributional shift between MCP-Atlas and {Who\&When}.

\begin{figure}
\floatsetup{heightadjust=all, valign=c}
\begin{floatrow}
\input{figures/cde_vs_rnn}
\input{figures/layer}
\end{floatrow}
\end{figure}

\paragraph{Comparison with RNN-based approach.}

We compare \approach with non-NDE one-class learning approach, particularly RNN, which directly takes $h_Q, h_1,\dots,h_{t-1}$ to predict $\hat{h}_t$. The anomaly score $e_t$ for RNN is defined as $e_t=\|h_t-\hat{h}_t\|_2^2$. 
Figure~\ref{fig:cde_vs_rnn} shows that \approach consistently outperforms RNN, suggesting that continuous model characterizes trajectory dynamics more accurately.

\paragraph{Representations at different layers.}

Rather than fixing representations to the last layer, we investigate how error step identification performance varies across model layers. Specifically, we evaluate representations extracted from layers $\{0, 8, 16, 24, 32, 40, 48, 56, 64\}$ of Qwen3.5-27B. Qwen3.5-27B has 64  layers; layer 0 corresponds to the embedding layer output. Figure~\ref{fig:layer} shows that later-layer representations consistently achieve higher precision, recall, and F1 score than earlier layers. This is consistent with the well-established finding in the representation learning literature that later layers of transformer models encode more abstract, semantic, and task-relevant information~\citep{gurnee2024language, 10.24963/ijcai.2025/566}, which provides a stronger signal for distinguishing normal from anomalous steps in trajectory modeling.

\paragraph{Additional ablations.}

We conduct other ablations on representation aggregation strategies, choice of proxy models, as well as the trade-off between precision and recall.
We also conduct additional experiments on the impact of training set size and CORAL.
See Appendix~\ref{ap:experiment} for more details.

\section{Conclusion}\label{sec:conclusion}

In this paper, we introduced unsupervised failure attribution, a new problem formulation for diagnosing failures in LLM-based agentic systems, eliminating the need for step-level annotation of failure trajectories. Rather than learning from labeled failure trajectories, we propose \approach, a Neural CDE-based one-class learning approach, training exclusively on successful trajectories and identifying error steps at inference time by detecting deviations from the learned dynamics of normal behavior. Experimental results show that \approach outperforms the prompting-based approach in both in-domain and OOD scenarios. Beyond performance, \approach runs 200--5000$\times$
faster than prompting-based approaches, making real-time failure attribution practical. We hope our work inspires a broader shift in how the community approaches agentic system diagnostics, leveraging the rich signal in successful trajectories for understanding, diagnosing, and ultimately preventing failure.

\section*{Acknowledgment}

We gratefully acknowledge Changdae Oh and Leitian Tao for their valuable comments on the draft. The work is supported by Microsoft Research. Sharon Li is also supported in part by the AFOSR
Young Investigator Program under award FA9550-23-1-0184, the Office of Naval Research 
under award N000142612508, the National Science Foundation under awards IIS-2237037 and
IIS-2331669, the Alfred P. Sloan Fellowship, Open Philanthropy (now Coefficient Giving), and
Schmidt Sciences Foundation.

\bibliographystyle{unsrtnat}
\bibliography{neurips_2026}

@inproceedings{3009657.3009740,
author = {Sch{\"o}lkopf, Bernhard and Williamson, Robert and Smola, Alex and Shawe-Taylor, John and Platt, John},
title = {Support vector method for novelty detection},
year = {1999},
booktitle = {Proceedings of the 13th International Conference on Neural Information Processing Systems},
}

@INPROCEEDINGS{4781136,
  author={Liu, Fei Tony and Ting, Kai Ming and Zhou, Zhi-Hua},
  booktitle={2008 Eighth IEEE International Conference on Data Mining}, 
  title={Isolation Forest}, 
  year={2008},
}

@inproceedings{
kong2026aegis,
title={Aegis: Automated Error Generation and Attribution for Multi-Agent Systems},
author={Fanqi Kong and Ruijie Zhang and Huaxiao Yin and Guibin Zhang and Xiaofei Zhang and Ziang Chen and Zhaowei Zhang and Xiaoyuan Zhang and Song-Chun Zhu and Xue Feng},
booktitle={The Fourteenth International Conference on Learning Representations},
year={2026},
}

@inproceedings{
zhang2025which,
title={Which Agent Causes Task Failures and When? On Automated Failure Attribution of {LLM} Multi-Agent Systems},
author={Shaokun Zhang and Ming Yin and Jieyu Zhang and Jiale Liu and Zhiguang Han and Jingyang Zhang and Beibin Li and Chi Wang and Huazheng Wang and Yiran Chen and Qingyun Wu},
booktitle={Forty-second International Conference on Machine Learning},
year={2025},
}

@inproceedings{
zhang2026agentracer,
title={AgenTracer: Who Is Inducing Failure in the {LLM} Agentic Systems?},
author={Guibin Zhang and Junhao Wang and Junjie Chen and Wangchunshu Zhou and Kun Wang and Shuicheng YAN},
booktitle={The Fourteenth International Conference on Learning Representations},
year={2026},
}

@inproceedings{
ma2026dover,
title={DoVer: Intervention-Driven Auto Debugging for {LLM} Multi-Agent Systems},
author={Ming Ma and Jue Zhang and Fangkai Yang and Yu Kang and Qingwei Lin and Saravan Rajmohan and Dongmei Zhang},
booktitle={The Fourteenth International Conference on Learning Representations},
year={2026},
}

@inproceedings{10.24963/ijcai.2025/1179,
author = {Oh, YongKyung and Kam, Seungsu and Lee, Jonghun and Lim, Dong-Young and Kim, Sungil and Bui, Alex A. T.},
title = {Comprehensive review of neural differential equations for time series analysis},
year = {2025},
booktitle = {Proceedings of the Thirty-Fourth International Joint Conference on Artificial Intelligence},
}

@inproceedings{10.5555/3454287.3454950,
author = {Brouwer, Edward De and Simm, Jaak and Arany, Adam and Moreau, Yves},
title = {GRU-ODE-Bayes: continuous modeling of sporadically-observed time series},
year = {2019},
booktitle = {Proceedings of the 33rd International Conference on Neural Information Processing Systems},
}

@inproceedings{NEURIPS2019_99a40143,
 author = {Yildiz, Cagatay and Heinonen, Markus and Lahdesmaki, Harri},
 booktitle = {Advances in Neural Information Processing Systems},
 title = {ODE2VAE: Deep generative second order ODEs with Bayesian neural networks},
 year = {2019}
}

@inproceedings{10.5555/3454287.3454765,
author = {Rubanova, Yulia and Chen, Ricky T. Q. and Duvenaud, David},
title = {Latent ODEs for irregularly-sampled time series},
year = {2019},
booktitle = {Proceedings of the 33rd International Conference on Neural Information Processing Systems},
}

@inproceedings{NEURIPS2018_69386f6b,
 author = {Chen, Ricky T. Q. and Rubanova, Yulia and Bettencourt, Jesse and Duvenaud, David K},
 booktitle = {Advances in Neural Information Processing Systems},
 title = {Neural Ordinary Differential Equations},
 year = {2018}
}

@inproceedings{10.5555/3495724.3496286,
author = {Kidger, Patrick and Morrill, James and Foster, James and Lyons, Terry},
title = {Neural controlled differential equations for irregular time series},
year = {2020},
booktitle = {Proceedings of the 34th International Conference on Neural Information Processing Systems},
}

@article{10.1561/2200000101,
author = {Angelopoulos, Anastasios N. and Bates, Stephen},
title = {Conformal Prediction: A Gentle Introduction},
year = {2023},
volume = {16},
number = {4},
issn = {1935-8237},
journal = {Found. Trends Mach. Learn.},
}

@article{10.1007/s10115-023-01977-5,
author = {Jhin, Sheo Yon and Shin, Heejoo and Kim, Sujie and Hong, Seoyoung and Jo, Minju and Park, Solhee and Park, Noseong and Lee, Seungbeom and Maeng, Hwiyoung and Jeon, Seungmin},
title = {Attentive neural controlled differential equations for time-series classification and forecasting},
year = {2023},
volume = {66},
number = {3},
issn = {0219-1377},
journal = {Knowl. Inf. Syst.},
}

@inproceedings{schmidgall-etal-2025-agent,
    title = "Agent Laboratory: Using {LLM} Agents as Research Assistants",
    author = "Schmidgall, Samuel  and
      Su, Yusheng  and
      Wang, Ze  and
      Sun, Ximeng  and
      Wu, Jialian  and
      Yu, Xiaodong  and
      Liu, Jiang  and
      Moor, Michael  and
      Liu, Zicheng  and
      Barsoum, Emad",
    booktitle = "Findings of the Association for Computational Linguistics: EMNLP 2025",
    year = "2025",
}

@article{lu2026towards,
  title={Towards end-to-end automation of AI research},
  author={Lu, Chris and Lu, Cong and Lange, Robert Tjarko and Yamada, Yutaro and Hu, Shengran and Foerster, Jakob and Ha, David and Clune, Jeff},
  journal={Nature},
  year={2026},
}

@inproceedings{
wang2025openhands,
title={OpenHands: An Open Platform for {AI} Software Developers as Generalist Agents},
author={Xingyao Wang and Boxuan Li and Yufan Song and Frank F. Xu and Xiangru Tang and Mingchen Zhuge and Jiayi Pan and Yueqi Song and Bowen Li and Jaskirat Singh and Hoang H. Tran and Fuqiang Li and Ren Ma and Mingzhang Zheng and Bill Qian and Yanjun Shao and Niklas Muennighoff and Yizhe Zhang and Binyuan Hui and Junyang Lin and Robert Brennan and Hao Peng and Heng Ji and Graham Neubig},
booktitle={The Thirteenth International Conference on Learning Representations},
year={2025},
}

@inproceedings{
jimenez2024swebench,
title={{SWE}-bench: Can Language Models Resolve Real-world Github Issues?},
author={Carlos E Jimenez and John Yang and Alexander Wettig and Shunyu Yao and Kexin Pei and Ofir Press and Karthik R Narasimhan},
booktitle={The Twelfth International Conference on Learning Representations},
year={2024},
}

@inproceedings{
li2023neural,
title={Neural Lad: A Neural Latent Dynamics Framework for Times Series Modeling},
author={Ting Li and Jianguo Li and Zhanxing Zhu},
booktitle={Thirty-seventh Conference on Neural Information Processing Systems},
year={2023},
}

@article{bandi2026mcpatlaslargescalebenchmarktooluse,
      title={MCP-Atlas: A Large-Scale Benchmark for Tool-Use Competency with Real MCP Servers}, 
      author={Chaithanya Bandi and Ben Hertzberg and Geobio Boo and Tejas Polakam and Jeff Da and Sami Hassaan and Manasi Sharma and Andrew Park and Ernesto Hernandez and Dan Rambado and Ivan Salazar and Rafael Cruz and Chetan Rane and Ben Levin and Brad Kenstler and Bing Liu},
      year={2026},
      journal={arXiv preprint arXiv:2602.00933},
}

@article{zhou2026rcnfrobotconditionednormalizingflow,
      title={RC-NF: Robot-Conditioned Normalizing Flow for Real-Time Anomaly Detection in Robotic Manipulation}, 
      author={Shijie Zhou and Bin Zhu and Jiarui Yang and Xiangyu Zhao and Jingjing Chen and Yu-Gang Jiang},
      year={2026},
      journal={arXiv preprint arXiv:2603.11106},
}

@inproceedings{
romer2026failure,
title={Failure Prediction at Runtime for Generative Robot Policies},
author={Ralf R{\"o}mer and Adrian Kobras and Luca Worbis and Angela P. Schoellig},
booktitle={The Thirty-ninth Annual Conference on Neural Information Processing Systems},
year={2026},
}

@inproceedings{
xu2025can,
title={Can We Detect Failures Without Failure Data? Uncertainty-Aware Runtime Failure Detection for Imitation Learning Policies},
author={Chen Xu and Tony Khuong Nguyen and Emma Dixon and Christopher Rodriguez and Patrick Miller and Robert Lee and Paarth Shah and Rares Andrei Ambrus and Haruki Nishimura and Masha Itkina},
booktitle={Robot Evaluation for the Real World},
year={2025},
}

@article{sun2016correlationalignmentunsuperviseddomain,
      title={Correlation Alignment for Unsupervised Domain Adaptation}, 
      author={Baochen Sun and Jiashi Feng and Kate Saenko},
      year={2016},
      journal={arXiv preprint arXiv:1612.01939},
}

@article{in2026rethinkingfailureattributionmultiagent,
      title={Rethinking Failure Attribution in Multi-Agent Systems: A Multi-Perspective Benchmark and Evaluation}, 
      author={Yeonjun In and Mehrab Tanjim and Jayakumar Subramanian and Sungchul Kim and Uttaran Bhattacharya and Wonjoong Kim and Sangwu Park and Somdeb Sarkhel and Chanyoung Park},
      year={2026},
      journal={arXiv preprint arXiv:2603.2500},
}

@article{chalapathy2019deeplearninganomalydetection,
      title={Deep Learning for Anomaly Detection: A Survey}, 
      author={Raghavendra Chalapathy and Sanjay Chawla},
      year={2019},
      journal={arXiv preprint arXiv:1901.03407},
}

@InProceedings{pmlr-v80-ruff18a,
  title = 	 {Deep One-Class Classification},
  author =       {Ruff, Lukas and Vandermeulen, Robert and Goernitz, Nico and Deecke, Lucas and Siddiqui, Shoaib Ahmed and Binder, Alexander and M{\"u}ller, Emmanuel and Kloft, Marius},
  booktitle = 	 {Proceedings of the 35th International Conference on Machine Learning},
  year = 	 {2018},
  volume = 	 {80},
}

@article{10.1023/B:MACH.0000008084.60811.49,
author = {Tax, David M. J. and Duin, Robert P. W.},
title = {Support Vector Data Description},
year = {2004},
volume = {54},
number = {1},
issn = {0885-6125},
journal = {Mach. Learn.},
}

@inproceedings{NIPS1999_8725fb77,
 author = {Sch\"{o}lkopf, Bernhard and Williamson, Robert C and Smola, Alex and Shawe-Taylor, John and Platt, John},
 booktitle = {Advances in Neural Information Processing Systems},
 title = {Support Vector Method for Novelty Detection},
 volume = {12},
 year = {1999}
}

@inproceedings{10.1145/2689746.2689747,
author = {Sakurada, Mayu and Yairi, Takehisa},
title = {Anomaly Detection Using Autoencoders with Nonlinear Dimensionality Reduction},
year = {2014},
isbn = {9781450331593},
booktitle = {Proceedings of the MLSDA 2014 2nd Workshop on Machine Learning for Sensory Data Analysis},
}

@inproceedings{10.1145/3292500.3330672,
author = {Su, Ya and Zhao, Youjian and Niu, Chenhao and Liu, Rong and Sun, Wei and Pei, Dan},
title = {Robust Anomaly Detection for Multivariate Time Series through Stochastic Recurrent Neural Network},
year = {2019},
isbn = {9781450362016},
booktitle = {Proceedings of the 25th ACM SIGKDD International Conference on Knowledge Discovery \& Data Mining},
}

@article{park2018multimodal,
  title={A multimodal anomaly detector for robot-assisted feeding using an lstm-based variational autoencoder},
  author={Park, Daehyung and Hoshi, Yuuna and Kemp, Charles C},
  journal={IEEE Robotics and Automation Letters},
  volume={3},
  number={3},
  year={2018},
}

@inproceedings{
xu2022anomaly,
title={Anomaly Transformer: Time Series Anomaly Detection with Association Discrepancy},
author={Jiehui Xu and Haixu Wu and Jianmin Wang and Mingsheng Long},
booktitle={International Conference on Learning Representations},
year={2022},
}

@article{Hayashi_2026,
   title={Critical Review for One‐Class Classification: Recent Advances and Reality Behind Them},
   volume={16},
   ISSN={1942-4795},
   number={1},
   journal={WIREs Data Mining and Knowledge Discovery},
   author={Hayashi, Toshitaka and Cimr, Dalibor and Fujita, Hamido and Cimler, Richard},
   year={2026},
   }

@article{10.1016/j.cose.2008.08.003,
author = {Garc\'{\i}a-Teodoro, P. and D\'{\i}az-Verdejo, J. and Maci\'{a}-Fern\'{a}ndez, G. and V\'{a}zquez, E.},
title = {Anomaly-based network intrusion detection: Techniques, systems and challenges},
year = {2009},
volume = {28},
number = {1–2},
issn = {0167-4048},
journal = {Comput. Secur.},
}

@article{10.1145/3691338,
author = {Zamanzadeh Darban, Zahra and Webb, Geoffrey I. and Pan, Shirui and Aggarwal, Charu and Salehi, Mahsa},
title = {Deep Learning for Time Series Anomaly Detection: A Survey},
year = {2024},
volume = {57},
number = {1},
issn = {0360-0300},
journal = {ACM Comput. Surv.},
}

@inproceedings{
yao2023react,
title={ReAct: Synergizing Reasoning and Acting in Language Models},
author={Shunyu Yao and Jeffrey Zhao and Dian Yu and Nan Du and Izhak Shafran and Karthik R Narasimhan and Yuan Cao},
booktitle={The Eleventh International Conference on Learning Representations },
year={2023},
}

@inproceedings{10.24963/ijcai.2024/890,
author = {Guo, Taicheng and Chen, Xiuying and Wang, Yaqi and Chang, Ruidi and Pei, Shichao and Chawla, Nitesh V. and Wiest, Olaf and Zhang, Xiangliang},
title = {Large language model based multi-agents: a survey of progress and challenges},
year = {2024},
booktitle = {Proceedings of the Thirty-Third International Joint Conference on Artificial Intelligence},
}

@misc{llama4,
  author       = {MetaAI},
  title        = {{Llama-4-Scout-17B-16E-Instruct}},
  year         = {2025},
  howpublished = {\url{https://huggingface.co/meta-llama/Llama-4-Scout-17B-16E-Instruct}}
}

@misc{gemma4,
  author       = {Google DeepMind},
  title        = {{Welcome Gemma 4: Frontier multimodal intelligence on device}},
  year         = {2026},
  howpublished = {\url{https://huggingface.co/blog/gemma4}}
}

@article{openai2025gptoss120bgptoss20bmodelv,
      title={gpt-oss-120b \& gpt-oss-20b Model Card}, 
      author={OpenAI},
      year={2025},
      journal={arXiv preprint arXiv:2508.10925}, 
}

@inproceedings{10.24963/ijcai.2025/566,
author = {Fan, Siqi and Jiang, Xin and Li, Xiang and Meng, Xuying and Han, Peng and Shang, Shuo and Sun, Aixin and Wang, Yequan},
title = {Not all layers of LLMs are necessary during inference},
year = {2025},
booktitle = {Proceedings of the Thirty-Fourth International Joint Conference on Artificial Intelligence},
}

@inproceedings{
gurnee2024language,
title={Language Models Represent Space and Time},
author={Wes Gurnee and Max Tegmark},
booktitle={The Twelfth International Conference on Learning Representations},
year={2024},
}

@article{Wang_2024,
   title={A survey on large language model based autonomous agents},
   volume={18},
   ISSN={2095-2236},
   number={6},
   journal={Frontiers of Computer Science},
   author={Wang, Lei and Ma, Chen and Feng, Xueyang and Zhang, Zeyu and Yang, Hao and Zhang, Jingsen and Chen, Zhiyuan and Tang, Jiakai and Chen, Xu and Lin, Yankai and Zhao, Wayne Xin and Wei, Zhewei and Wen, Jirong},
   year={2024}}

@article{qian2026actionunveilinginternaldrivers,
      title={The Why Behind the Action: Unveiling Internal Drivers via Agentic Attribution}, 
      author={Chen Qian and Peng Wang and Dongrui Liu and Junyao Yang and Dadi Guo and Ling Tang and Jilin Mei and Qihan Ren and Shuai Shao and Yong Liu and Jie Fu and Jing Shao and Xia Hu},
      year={2026},
      journal={arXiv preprint arXiv:2601.15075}, 
}

@article{morrill2021neuralcontrolleddifferentialequations,
      title={Neural Controlled Differential Equations for Online Prediction Tasks}, 
      author={James Morrill and Patrick Kidger and Lingyi Yang and Terry Lyons},
      year={2021},
      journal={arXiv preprint arXiv:2106.11028}, 
}

@article{tzen2019neuralstochasticdifferentialequations,
      title={Neural Stochastic Differential Equations: Deep Latent Gaussian Models in the Diffusion Limit}, 
      author={Belinda Tzen and Maxim Raginsky},
      year={2019},
      journal={arXiv preprint arXiv:1905.09883},  
}

@article{liu2025largelanguagemodelbasedagents,
author = {Liu, Junwei and Wang, Kaixin and Chen, Yixuan and Peng, Xin and Chen, Zhenpeng and Zhang, Lingming and Lou, Yiling},
title = {Large Language Model-Based Agents for Software Engineering: A Survey},
year = {2026},
issn = {1049-331X},
journal = {ACM Trans. Softw. Eng. Methodol.},
}

@article{ren2026scientificintelligencesurveyllmbased,
      title={Towards Scientific Intelligence: A Survey of LLM-based Scientific Agents}, 
      author={Shuo Ren and Can Xie and Pu Jian and Zhenjiang Ren and Chunlin Leng and Jiajun Zhang},
      year={2026},
      journal={arXiv preprint arXiv:2503.24047}, 
}

@article{song2025adaptiveinconversationteambuilding,
      title={Adaptive In-conversation Team Building for Language Model Agents}, 
      author={Linxin Song and Jiale Liu and Jieyu Zhang and Shaokun Zhang and Ao Luo and Shijian Wang and Qingyun Wu and Chi Wang},
      year={2025},
      journal={arXiv preprint arXiv:2405.19425}
}

@article{fourney2024magenticonegeneralistmultiagentsolving,
      title={Magentic-One: A Generalist Multi-Agent System for Solving Complex Tasks}, 
      author={Adam Fourney and Gagan Bansal and Hussein Mozannar and Cheng Tan and Eduardo Salinas and Erkang and Zhu and Friederike Niedtner and Grace Proebsting and Griffin Bassman and Jack Gerrits and Jacob Alber and Peter Chang and Ricky Loynd and Robert West and Victor Dibia and Ahmed Awadallah and Ece Kamar and Rafah Hosn and Saleema Amershi},
      year={2024},
      journal={arXiv preprint arXiv:2411.04468}
}

@inproceedings{
mialon2024gaia,
title={{GAIA}: a benchmark for General {AI} Assistants},
author={Gr{\'e}goire Mialon and Cl{\'e}mentine Fourrier and Thomas Wolf and Yann LeCun and Thomas Scialom},
booktitle={The Twelfth International Conference on Learning Representations},
year={2024},
}

@inproceedings{yoran-etal-2024-assistantbench,
    title = "{A}ssistant{B}ench: Can Web Agents Solve Realistic and Time-Consuming Tasks?",
    author = "Yoran, Ori  and
      Amouyal, Samuel Joseph  and
      Malaviya, Chaitanya  and
      Bogin, Ben  and
      Press, Ofir  and
      Berant, Jonathan",
    booktitle = "Proceedings of the 2024 Conference on Empirical Methods in Natural Language Processing",
    year = "2024",
}

@article{deshpande2025trailtracereasoningagentic,
      title={TRAIL: Trace Reasoning and Agentic Issue Localization}, 
      author={Darshan Deshpande and Varun Gangal and Hersh Mehta and Jitin Krishnan and Anand Kannappan and Rebecca Qian},
      year={2025},
      journal={arXiv preprint arXiv:2505.08638},
}

@inproceedings{
cemri2025why,
title={Why Do Multi-Agent {LLM} Systems Fail?},
author={Mert Cemri and Melissa Z Pan and Shuyi Yang and Lakshya A Agrawal and Bhavya Chopra and Rishabh Tiwari and Kurt Keutzer and Aditya Parameswaran and Dan Klein and Kannan Ramchandran and Matei Zaharia and Joseph E. Gonzalez and Ion Stoica},
booktitle={The Thirty-ninth Annual Conference on Neural Information Processing Systems Datasets and Benchmarks Track},
year={2025},
}

@inproceedings{
banerjee2025didwronghierarchicallook,
title={Where Did It All Go Wrong? A Hierarchical Look into Multi-Agent Error Attribution},
author={Adi Banerjee and Anirudh Nair and Tarik Borogovac},
booktitle={NeurIPS 2025 Workshop on Evaluating the Evolving LLM Lifecycle: Benchmarks, Emergent Abilities, and Scaling},
year={2025},
}

@article{wang2026flatlogscausalgraphs,
      title={From Flat Logs to Causal Graphs: Hierarchical Failure Attribution for LLM-based Multi-Agent Systems}, 
      author={Yawen Wang and Wenjie Wu and Junjie Wang and Qing Wang},
      year={2026},
      journal={arXiv preprint arXiv:2602.23701},
}

@article{zhang2025graphtracergraphguidedfailuretracing,
      title={GraphTracer: Graph-Guided Failure Tracing in LLM Agents for Robust Multi-Turn Deep Search}, 
      author={Heng Zhang and Yuling Shi and Xiaodong Gu and Haochen You and Zijian Zhang and Lubin Gan and Yilei Yuan and Jin Huang},
      year={2025},
      journal={arXiv preprint arXiv:2510.10581},
}

@article{shao2024deepseekmathpushinglimitsmathematical,
      title={DeepSeekMath: Pushing the Limits of Mathematical Reasoning in Open Language Models}, 
      author={Zhihong Shao and Peiyi Wang and Qihao Zhu and Runxin Xu and Junxiao Song and Xiao Bi and Haowei Zhang and Mingchuan Zhang and Y. K. Li and Y. Wu and Daya Guo},
      year={2024},
      journal={arXiv preprint arXiv:2402.03300},
}

@misc{qwen3.5,
    title  = {{Qwen3.5}: Towards Native Multimodal Agents},
    author = {{Qwen Team}},
    month  = {February},
    year   = {2026},
    url    = {https://qwen.ai/blog?id=qwen3.5}
}

@article{openai2024gpt4ocard,
      title={GPT-4o System Card}, 
      author={OpenAI },
      year={2024},
      journal={arXiv preprint arXiv:2410.21276},
}

@article{singh2025openaigpt5card,
      title={OpenAI GPT-5 System Card}, 
      author={OpenAI},
      year={2025},
      journal={arXiv preprint arXiv:2601.03267},
}

\newpage
\appendix

\textsc{\huge {Appendix}}

\addcontentsline{toc}{section}{Appendix}

\startcontents[appendix]

\vspace{1.5em}
\textsc{\Large Contents}

\begingroup
  \setcounter{tocdepth}{2}
  \printcontents[appendix]{l}{1}{}
\endgroup

\section{Limitations and future work}\label{ap:limitation}

\approach identifies failure-contributing steps via anomaly detection, which is not equivalent to causal attribution. Thus, \approach cannot guarantee that an identified step is causally failure-contributing. While not every anomaly is causal and not every causal error must appear anomalous, our empirical results in Section~\ref{sec:experiment} and qualitative analyses in Appendix~\ref{ap:case_study} demonstrate that deviations from successful trajectory dynamics provide an effective surrogate signal for identifying likely failure-contributing steps in practice across both in-domain and out-of-distribution benchmarks.

In this work, we explore the feasibility of modeling successful trajectories with NDEs and demonstrate that Neural CDEs are effective for continuous-time modeling of agent trajectories. Beyond vanilla NDE, there are many extensions that can be explored. For example,  \citet{NEURIPS2018_69386f6b} showed that NDEs can be extended as Continuous Normalizing Flows (CNFs), enabling exact likelihood computation via the instantaneous change-of-variables formula. Under this formulation, the anomaly score for each step could be replaced by its negative log-likelihood under the learned distribution, providing a more principled measure of deviation from normal behavior. Alternatively, one can incorporate Neural CDEs with attention mechanisms for control path construction~\citep{10.1007/s10115-023-01977-5, li2023neural}, which may be beneficial for distinguishing between decisive and trivial steps and better trajectory modeling. Beyond density estimation, Neural CDEs can also be extended with stochastic dynamics via Neural Stochastic Differential Equations (Neural SDEs)~\citep{tzen2019neuralstochasticdifferentialequations}, which model uncertainty in the latent trajectory evolution and may better capture the inherent stochasticity of LLM-generated trajectories. These extensions suggest that there is a huge room to explore in unsupervised failure attribution. We leave such exploration to future work.

Another limitation of \approach is that it requires training on a small set of successful trajectories. Although, compared to step-level failure annotations, successful trajectories are substantially less burdensome to collect and often available from deployment logs, demonstrations, or automated testing in practice, we acknowledge that there could be scenarios where successful trajectories are rare or hard to define. Thus, future directions on unsupervised failure attribution without any labeled trajectories would be worth exploring.

\section{Societal Impact}\label{ap:societal_impact}

This work aims to improve the transparency and reliability of LLM-based agentic systems by enabling automatic identification of error contributing steps without human annotation. As agentic systems are increasingly deployed in high-stakes domains such as software engineering, scientific research, and automated decision making, the ability to identify and explain failures is an important step toward building safer systems. By reducing the annotation burden for failure attribution, our approach also lowers the barrier of monitoring and debugging deployed agents, which may encourage more responsible deployment practices. We do not foresee direct negative societal impacts of this work. However, we note that failure attribution tools could be used to optimize agents for evading detection rather than for genuine improvement, and that automated failure attribution should be treated as an aid to human oversight rather than a replacement for it.

\section{Reproducibility Statement}\label{ap:reproducibility}

We detail the implementation of \approach in Section~\ref{sec:setup} and Appendix~\ref{ap:implementation}, including representation extraction, model structure of neural CDE vector field, CDE solver, construction of control path, as well as all corresponding hyperparameters. In Section~\ref{sec:setup}, we also illustrate the experimental settings, including baselines, datasets, and evaluation metrics. The details of datasets and step-level error annotation are further provided in Appendix~\ref{ap:dataset} and~\ref{ap:annotation}.  These comprehensive reports will help future studies reproduce our experiments. Code and data are available at \url{https://github.com/deeplearning-wisc/OAT}.

\begin{table}[t]
    \centering
    \small
    \resizebox{\textwidth}{!}{
    \begin{tabular}{lccccccc}
        \toprule
        & \# Samples & Avg. \# Steps & Max \# Steps & Avg. \# Tokens & Max \# Tokens & Avg. \# Errors & Max \# Errors\\
        \midrule
         Successful & 103 & 7.18 & 19 & 3.6K & 18K & - & -\\
         Failure & 88 & 7.32 & 20 & 4.2K & 20K & 1.56 & 12\\
         \bottomrule
    \end{tabular}
    }
    \caption{\textbf{Statistics of MCP-Atlas dataset.}}
    \label{tb:mcp_atlas_stat}
\end{table}

\section{Details of Datasets}\label{ap:dataset}

\paragraph{MCP-Atlas.} MCP-Atlas~\citep{bandi2026mcpatlaslargescalebenchmarktooluse} is a benchmark for evaluating tool-use agents, covering 36 MCP servers and 220 tools across 1,000 tasks designed to assess tool-use competency in realistic, multi-step workflows. Agents are evaluated via a claims-based rubric using LLM-as-a-Judge. In our experiments, we run Qwen3.5-27B on 203 tasks drawn from the public subset for which we have free access to the required MCP server APIs. 
Each trajectory is sampled directly through vLLM with the default generation config of Qwen3.5 (temperature=0.6, top\_k=20, top\_p=0.95). 
We use the same Qwen3.5-27B as judge to evaluate each trajectory against the claims-based rubric, and manually verify all judging results. A trajectory is considered successful if it satisfies all rubric claims, and a failure if it violates at least one. This yields 103 successful and 100 failure trajectories. The successful trajectories form our training set, while the failure trajectories form our test set. For the failure trajectories, we filter out 12 trajectories, in which the failure is not caused by agent behavior. We then manually annotate step-level failure-contributing steps for the rest of 88 failure trajectories; the annotation procedure is detailed in Appendix~\ref{ap:annotation}. Table~\ref{tb:mcp_atlas_stat} summarizes the trajectory statistics for MCP-Atlas.

\paragraph{Who\&When.} Who\&When~\citep{zhang2025which} is a failure attribution benchmark for LLM-based multi-agent systems. The tasks are drawn from two sources: GAIA~\citep{mialon2024gaia}, which contains queries requiring diverse modality processing (\eg, PDFs, spreadsheets, images, videos, and audio), web browsing, and coding; and AssistantBench~\citep{yoran-etal-2024-assistantbench}, which requires agents to interact with multiple websites across topics in biology, geography, and visual arts. Trajectories were generated by two agentic systems, CaptainAgent~\citep{song2025adaptiveinconversationteambuilding} and Magnetic-One~\citep{fourney2024magenticonegeneralistmultiagentsolving}, both using GPT-4o as the base model. The dataset consists entirely of failure trajectories, comprising 184 trajectories in total, each annotated with a step-level label identifying the first decisive error. Since Who\&When contains only failure trajectories, we use it exclusively as a test set, evaluating our model trained on MCP-Atlas successful trajectories. Table~\ref{tb:who_and_when_stat} summarizes the trajectory statistics for Who\&When.

\begin{table}[t]
    \centering
    \small
    
    \begin{tabular}{ccccc}
        \toprule
        \# Samples & Avg. \# Steps & Max \# Steps & Avg. \# Tokens & Max \# Tokens \\
        \midrule
         184 & 20.28 & 129 & 6.8K & 60K\\
         \bottomrule
    \end{tabular}
    
    \caption{\textbf{Statistics of Who\&When dataset.}}
    \label{tb:who_and_when_stat}
\end{table}

\section{Annotation of Failure-Contributing Steps}\label{ap:annotation}

\begin{figure}[t]
    \centering
    \includegraphics[width=\textwidth]{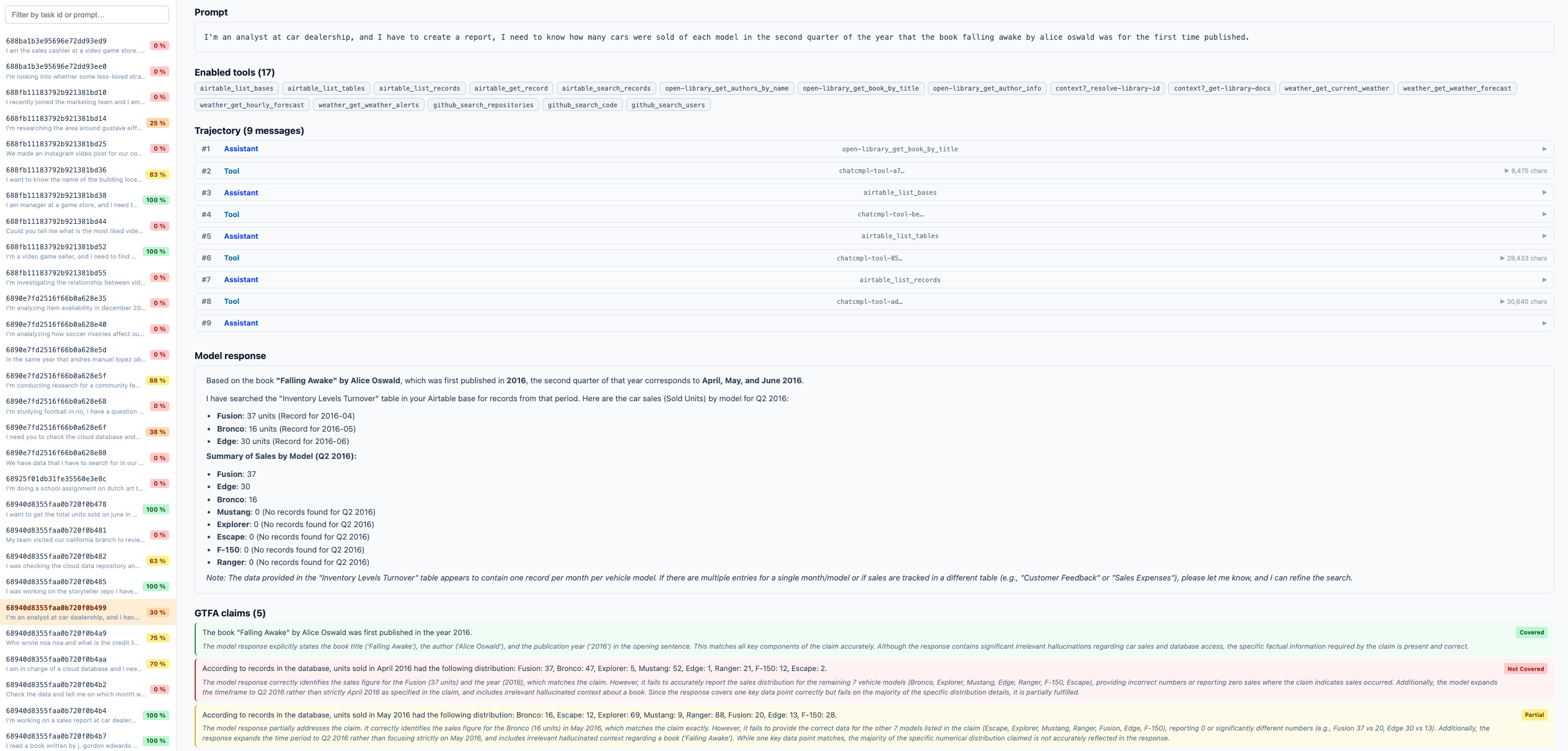}
    \caption{\textbf{Visualization tool for trajectory annotation.} We develop a tool to visualize trajectories and help tracing errors. The tool shows the complete trajectory as well as the claim-level judging results.}
    \label{fig:annotation_system}
\end{figure}

For each failure trajectory in MCP-Atlas, we manually annotate the set of failure-contributing steps $C(\tau)$ following the definition in Section~\ref{sec:problem_statement}. Figure~\ref{fig:annotation_system} shows the visualization tool for annotation. The annotation process proceeds in two stages: trajectory filtering and step-level labeling.

\paragraph{Trajectory Filtering.}

Prior to annotation, we exclude trajectories whose failure is attributable to infrastructure errors rather than agent behavior. Specifically, we identified a particular search API that consistently returns timeout errors regardless of the query or context. Failures caused purely by this API error are not informative for evaluating failure attribution methods, as the contributing step is trivially identifiable and reflects an environmental rather than an agent-level fault. We therefore exclude all trajectories where the sole cause of failure is this timeout issue. We also exclude cases that the trajectory logs are empty.

\paragraph{Step-Level Annotation.}
For each retained failure trajectory, we use the claim-level judging results produced during evaluation as a guide to understand the nature and location of the failure. We then read through the full trajectory step by step and label a step as a contributing step if it exhibits one or more of the following error types:

\begin{itemize}[leftmargin=15pt]
    \item \textbf{Hallucination in reasoning content}: the agent produces factually incorrect or fabricated information in its reasoning, leading to a misguided subsequent action or incorrect planning.
    \item \textbf{Wrong arguments in tool-calling}: the agent invokes a tool with incorrect, malformed, or semantically inappropriate arguments that cause the tool call to fail or return an incorrect result.
    \item \textbf{Error messages in tool returns}: the tool returns an explicit error or exception that the agent fails to handle appropriately, leading to downstream failure.
\end{itemize}

\paragraph{Treatment of Exploratory Tool Use.}
A special consideration arises for search-type tool calls, where the agent may issue multiple queries with different keywords in an attempt to retrieve relevant information. We treat such behavior as a natural exploration process and do not penalize individual failed search attempts, as trying alternative phrasings is a reasonable and expected strategy. However, if the agent issues the same query (or a semantically equivalent one) multiple times despite receiving the same unsuccessful result, we interpret this as a failure to adapt and mark all repeated attempts after the first as failure-contributing steps. The first attempt is not marked as an error, as it represents legitimate exploratory behavior.

\paragraph{Annotation Quality.}
All annotations were performed by the authors, with each trajectory reviewed independently. Trajectories with ambiguous error attribution (\eg, where a failure could be attributed to either a reasoning error or an uninformative tool return) were discussed until consensus was reached. The final contributing step labels reflect the annotators' best judgment of which steps meaningfully degraded the trajectory toward failure.

\begin{table}[t]
    \centering
    \small
    \begin{tabular}{lc}
        \toprule
        Hyperparameter & Value \\
        \midrule
        Representation Extraction\\
        \quad Aggregation & Mean Pooling\\
        \quad Layer & Last \\
        \quad PCA Dim & 64\\
        \midrule
        Neural CDE\\
        \quad Hidden Dim & 64\\
        \quad \# Hidden Layer & 3\\
        \quad Control Path &  Cubic Spline\\
        \quad Gate Hidden Dim & 12\\
        \quad \# Gate Hidden Layer & 4\\
        \midrule
        Detection\\
        \quad $k$ (Top-$k$) & 3\\
        \quad $\alpha$ (CP) & 0.2\\
        \midrule
        Training\\
        \quad Learning Rate & $4e^{-5}$\\
        \quad  Batch Size & $32$\\
        \quad  Epoch & $300$\\
        \quad  Weight Decay & $1e^{-5}$\\
        \bottomrule
    \end{tabular}
    \caption{\textbf{Hyperparameters of \approach.}}
    \label{tb:hyperparameters}
\end{table}

\section{Additional Implementation Details \& Hyperparameters}\label{ap:implementation}

Section~\ref{sec:setup} introduces the core implementation details of our approach. Here we provide a comprehensive summary of all hyperparameters used in our experiments in Table~\ref{tb:hyperparameters}. Key design choices are ablated and discussed in Section~\ref{sec:ablation} and Appendix~\ref{ap:experiment}, covering representation extraction strategies, the use of Neural CDE with gated control path, and the selection of $k$ for top-$k$ detection and $\alpha$ for conformal prediction detection.

\section{Additional Experimental Results}\label{ap:experiment}

\subsection{Precision-Recall Trade-off}

In Section~\ref{sec:score}, we introduce top-$k$ detection and conformal prediction detection, where both of them have a hyperparameter (\ie, $k$ for top-$k$ detection and $\alpha$ for conformal prediction detection) that control the trade-off between precision and recall. Figure~\ref{fig:diff_k_alpha} shows the ablation study on $k\in\{1, 3, 5\}$ and $\alpha\in\{0.05, 0.1, 0.2\}$. The result shows that \approach trades off precision with recall as $k$ and $\alpha$ increase, whereas F1 score remains stable. In practice, the hyperparameters $k$ and $\alpha$ can be choose depending on whether the system requires a high precision or recall.

\begin{figure}[t]
\floatsetup{heightadjust=all, valign=c}
\begin{floatrow}
\input{figures/diff_k_alpha}
\input{figures/last_vs_mean}
\end{floatrow}
\end{figure}

\subsection{Ablations on Step Representations}\label{ap:ablation}

We ablate on design choices corresponding to step representation extraction, including representation aggregation strategies and the proxy LLM setting. All of these ablation studies were conducted on MCP-Atlas conformal prediction detection.

\paragraph{Mean pooling vs.\ last token.}

In Section~\ref{sec:setup}, we obtain step representations by applying mean pooling to aggregate token representations within each step. We ablate this choice by replacing mean pooling with the last token representation. Figure~\ref{fig:last_vs_mean} shows that this substitution significantly degrades performance. This suggests that the last token alone is insufficient to encode the complexity of an action step. Error signals in agent behavior often manifest in intermediate tokens, for instance, in the reasoning content or argument construction within a step, rather than at the final token. Mean pooling aggregates information across all tokens within a step, capturing a richer and more representative encoding of the step's content, which in turn improves the quality of the modeled trajectory.

\begin{figure}[t]
    \centering
    \includegraphics[width=\linewidth]{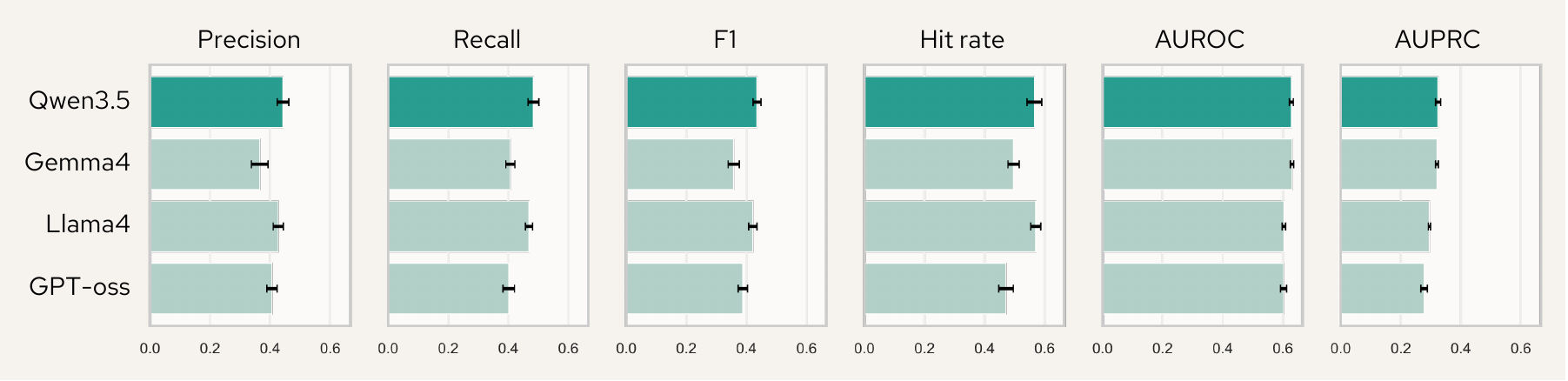}
    \caption{\textbf{Performance is stable under the proxy LLM setting.}}
    \label{fig:proxy_model}
\end{figure}

\paragraph{Representations extracted by different proxy LLMs.}

Instead of using the same Qwen3.5-27B model to extract representations of its own trajectories, we explore the possibility of proxy LLM setup, \ie, extracting representations with LLMs different from the one for generation. We examine three different LLMs, including 
Llama-4-Scout~\citep{llama4}, Gemma-4-31B~\citep{gemma4}, and GPT-oss-120B~\citep{openai2025gptoss120bgptoss20bmodelv}. 
Specifically, we extract trajectory representations from Qwen3.5-generated trajectories using different encoder models, and then train and evaluate \approach using these new representations. 
Figure~\ref{fig:proxy_model} shows that \approach works well under the proxy setting, with only a small performance degradation compared to the same-LLM setting. This result highlights the generalizability of \approach, enabling it to be applied in different practical scenarios where the generator LLM is not accessible.

\subsection{Localizing Onset Error}

Beyond the 6 metrics, we additionally analyze the ability of \approach to localize onset error. To do so, we compute the percentage of the earliest step predicted by \approach~(CP) that falls before and after the annotated first decisive error on Who\&When. We found that the percentages of falling before/after the first decisive error are 68.13\%/31.87\%, suggesting that \approach localizes onset 2 times often as downstream symptoms.
We also compute the percentages for the top-1 step scored by \approach, yielding 50\%/50\%, suggesting that \approach identifies half of onsets as steps with the largest deviation from successful dynamics.

\begin{figure}[t]
\floatsetup{heightadjust=all, valign=c}
\begin{floatrow}
\input{figures/training_set_size}
\input{figures/coral}
\end{floatrow}
\end{figure}

\subsection{Effect of the Size of Training Set}

\approach requires training on a small set of successful trajectories. To understand how many trajectories are needed, we sample 25, 50, 75, or 90 trajectories to train \approach. We run each setting 5 times to reduce the impact of randomness. Figure~\ref{fig:training_set_size} shows that a larger sample set improves performance. The result also shows that the standard deviations across 5 runs are small after a sample size $> 50$, suggesting that OAT is not highly sensitive to sampled trajectories.

\subsection{Effect of CORAL on OOD Setting}

In the OOD evaluation in Section~\ref{sec:result}, we fit CORAL on the entire Who\&When dataset.
To ensure that the representation of each evaluated trajectory is not used when estimating its own target-domain statistics, we additionally perform five-fold cross-validation. For each fold, CORAL statistics are estimated from the remaining four unlabeled folds, and evaluation is conducted on the held-out fold. Figure~\ref{fig:coral} shows the results. \approach with 5-fold CORAL underperforms its full CORAL counterpart. This is expected because the evaluated trajectory no longer informs its own alignment, and the statistics are estimated from only 80\% of the target data, making the covariance estimates of high-dimensional representations noisier. Nevertheless, 5-fold CORAL remains competitive with GPT-5, the strongest baseline. These results indicate that representation shift is a critical factor in OOD performance, and that the benefit of CORAL does not depend on including the evaluated trajectory in the target-domain statistics.

\section{Case Studies}\label{ap:case_study}

We randomly select three trajectories from MCP-Atlas where \approach identifies at least one annotated contributing step (\emph{successful cases}), and three trajectories where it fails to identify any annotated step (\emph{failure cases}). We analyze these cases to understand the conditions under which \approach succeeds and fails. We use conformal prediction detection for all case study outputs.

\subsection{Successful Cases}

\paragraph{Case 1.}

Figure~\ref{fig:success_1} in Section~\ref{sec:experiment} shows a task requiring the agent to find the oldest hospital in the US and nearby electric vehicle charging stations. At step 6, the agent attempts to retrieve information about the oldest hospital via the Wikipedia API but fails to obtain a result. At step 7, the agent falls back on its parametric knowledge and hallucinates an incorrect answer. Based on this fabricated result, the agent searches for nearby charging stations at step 8. Although the search itself succeeds, the query is wrong, and the error propagates to the final step, yielding an incorrect answer.

\approach assigns a significantly high anomaly score to the hallucinated step 7 while keeping the scores of benign steps low. Notably, the final step is also flagged as a contributing step, but with a more moderate score, reflecting that its logic is internally coherent while being contingent on the erroneous prior step. This case demonstrates that \approach is sensitive to the origin of an error and appropriately attenuates scores for downstream steps that inherit rather than introduce errors.

\begin{figure}
    \centering
    \includegraphics[width=\linewidth]{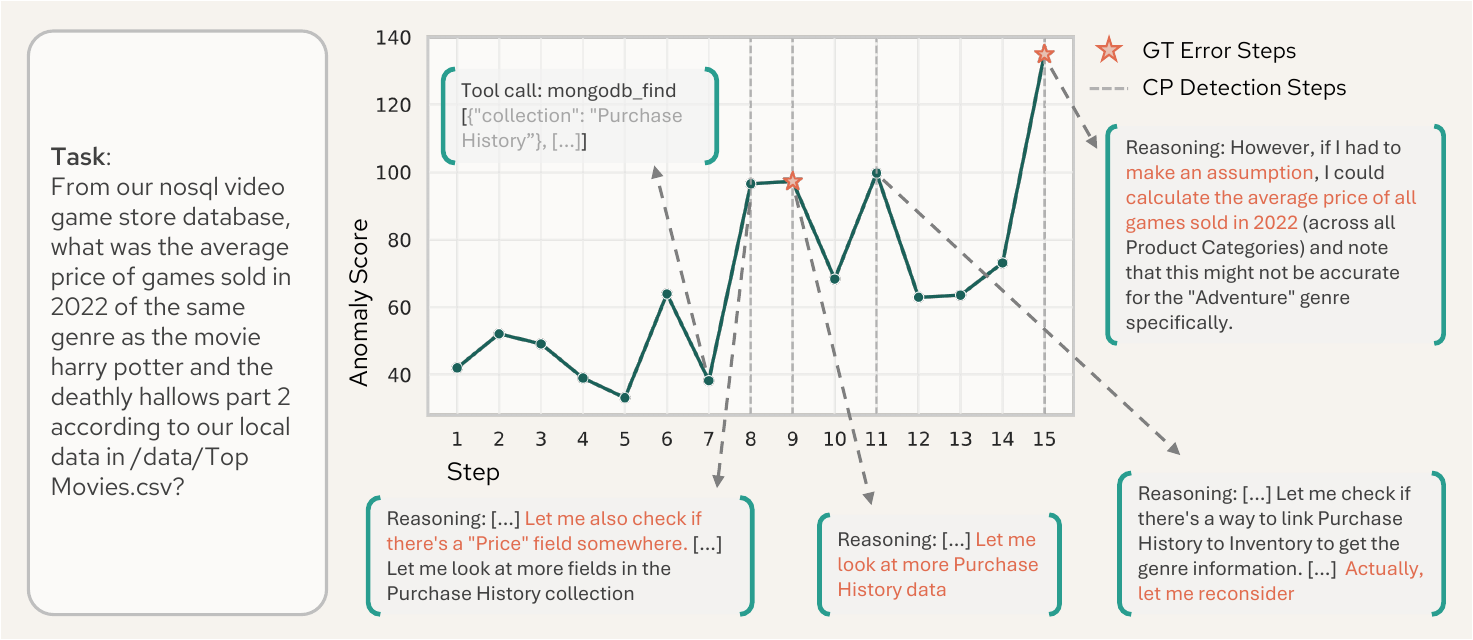}
    \caption{\textbf{Second successful case.} \approach identifies repeated useless actions at step 9, inconsistent reasoning at steps 8 and 11, and an unfaithful assumption at step 15.}
    \label{fig:success_2}
\end{figure}

\paragraph{Case 2.}

Figure~\ref{fig:success_2} shows a data analysis task on a database. At step 7, the agent calls a tool to fetch information and discovers that the target collection does not contain the required data. Over steps 8--14, the agent attempts various approaches to retrieve the information, exhibiting inconsistent reasoning\footnote{Because the agent still calls a reasonable tool at the end of these steps, we do not label the inconsistent reasoning itself as a contributing step.} and repeatedly re-attempting actions that had already been shown to be ineffective. At step 15, the agent resorts to an unfaithful assumption to work around its failure to retrieve the required information, ultimately producing an incorrect answer.

\approach assigns high anomaly scores to both annotated contributing steps, with step 15 receiving a particularly elevated score reflecting the severity of the unfaithful assumption and its direct causal role in the failure. \approach also flags steps 8 and 11, where the reasoning traces are internally inconsistent, despite these steps not being explicitly annotated as contributing steps. This behavior suggests that \approach captures subtle precursors to failure that may not meet the threshold for annotation but nonetheless indicate degraded trajectory dynamics. This case further illustrates that the first error does not necessarily receive the highest anomaly score: a later, more consequential error can dominate the scoring.

\begin{figure}
    \centering
    \includegraphics[width=\linewidth]{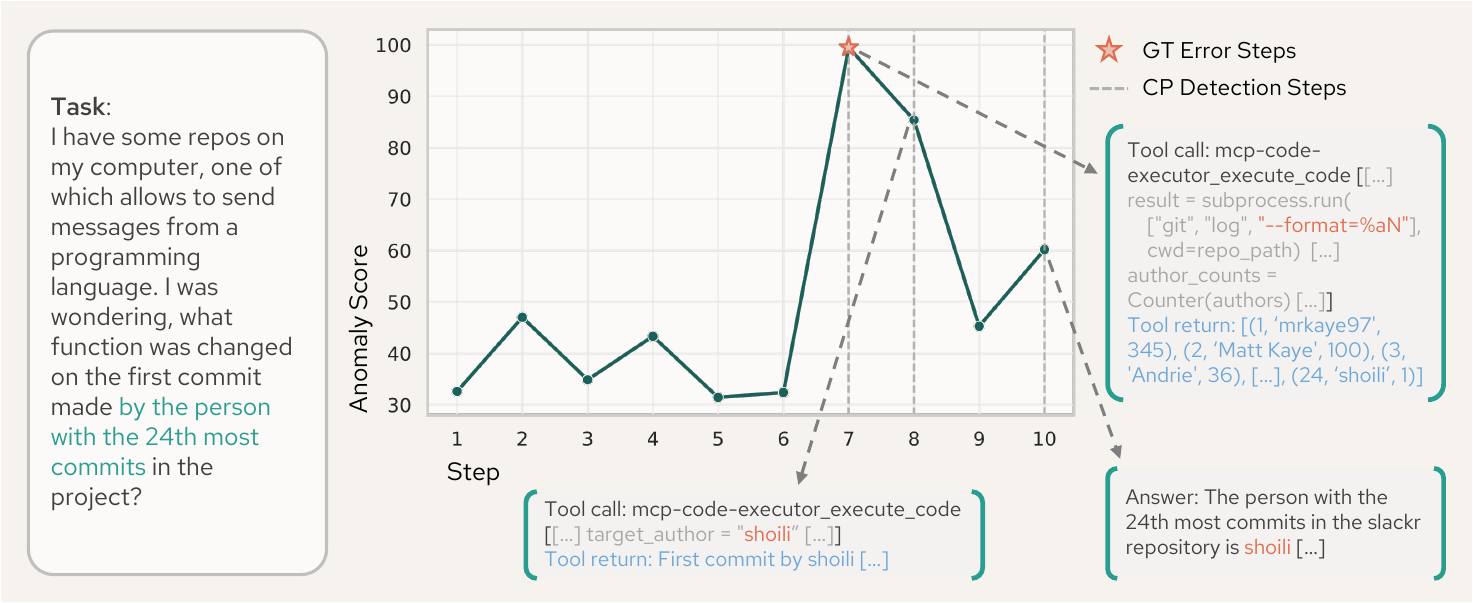}
    \caption{\textbf{Third successful case.} \approach identifies the bug at step 7 and the propagated errors at steps 8 and 10.}
    \label{fig:success_3}
\end{figure}

\paragraph{Case 3.}

Figure~\ref{fig:success_3} shows a task requiring the agent to find the user with the 24th most comments on a project. At step 7, the agent counts comments by display name rather than email address, accidentally merging two distinct users who share the same name. As a result, the code returns an incorrect user. This error propagates to step 8, where the agent writes a downstream code block based on the wrong result, and to step 10, where the incorrect answer is returned.

\approach assigns a significantly high anomaly score to the erroneous step 7 while keeping scores for benign steps low. The scores for steps 8 and 10 are elevated but progressively lower than step 7, consistent with the interpretation that these steps carry forward an inherited error rather than introducing a new one. This monotonic decay in anomaly scores along the error propagation chain is an encouraging property of the model's learned dynamics.

\begin{prompt}
\textbf{Summary.}

\medskip
Across the three successful cases, we found that \approach consistently assigns significantly high anomaly scores to steps that introduce errors, including  hallucination, unfaithful assumptions, or logical bugs, while keeping scores for benign steps low. In addition, when errors propagate downstream, anomaly scores decay gradually, reflecting that inherited errors are less anomalous than the steps that originate them. Notably, \approach also surfaces latent signals at steps that are not explicitly annotated as failure-contributing steps, such as internally inconsistent reasoning traces, suggesting that its learned notion of trajectory normality is sensitive to subtle precursors of failure beyond the ground truth labels.
\end{prompt}

\subsection{Failure Cases}

\begin{figure}
    \centering
    \includegraphics[width=\linewidth]{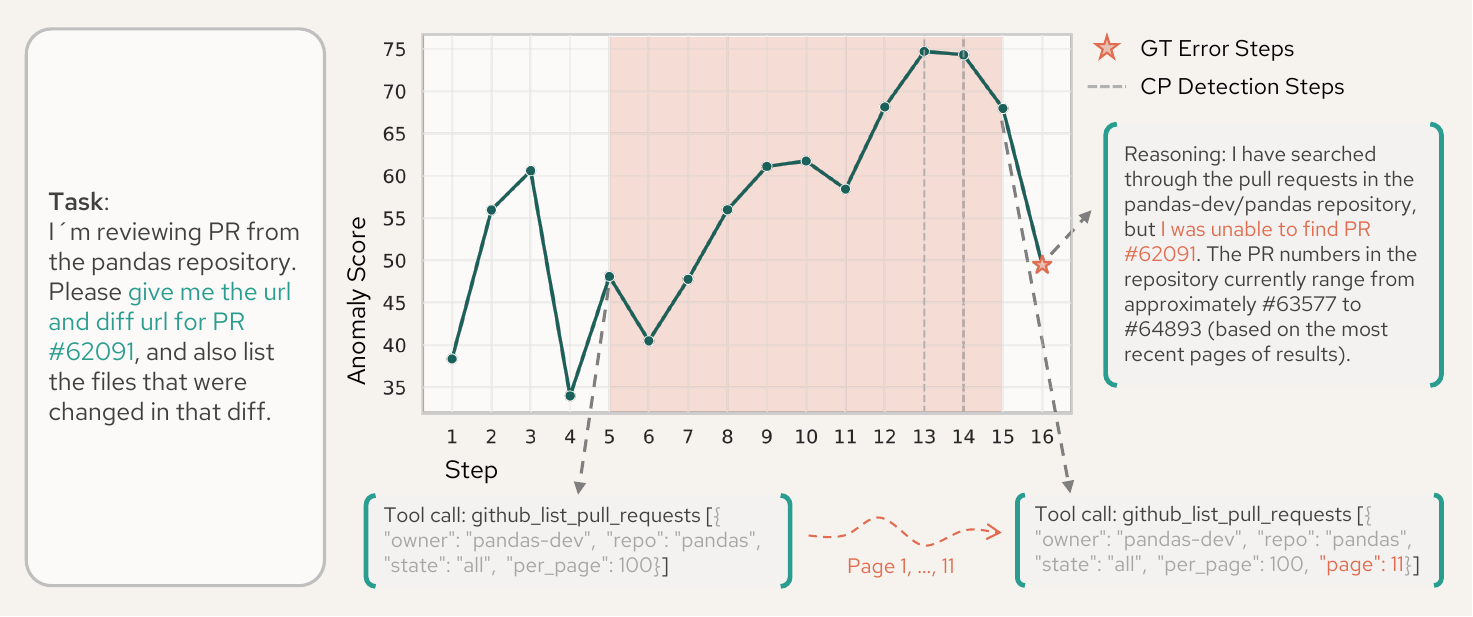}
    \caption{\textbf{First failure case.} \approach does not identify the incorrect answer at step 16 but assigns progressively increasing anomaly scores to the suboptimal plan spanning steps 5--15.}
    \label{fig:fail_1}
\end{figure}

\paragraph{Case 1.}

Figure~\ref{fig:fail_1} shows a task requiring the agent to retrieve information about a specific pull request. The agent adopts a suboptimal strategy of paginating through all pull requests sequentially from step 5 to step 15. While this approach is not inherently incorrect, it is inefficient and leads to step 16, where the agent incorrectly concludes that the target PR could not be found.

\approach does not flag step 16 as a contributing step, likely because the verbalization of an inability to find a result is not inherently anomalous from the model's perspective. (A successful trajectory could plausibly contain a similar statement in a different context.) Nevertheless, \approach assigns progressively increasing anomaly scores across the suboptimal steps 5--15, suggesting that it recognizes the suboptimal plan as increasingly deviant from the dynamics of successful trajectories, even in the absence of an explicit error signal. This case highlights that \approach can surface latent failure precursors that are not captured by the ground truth annotation.

\begin{figure}
    \centering
    \includegraphics[width=\linewidth]{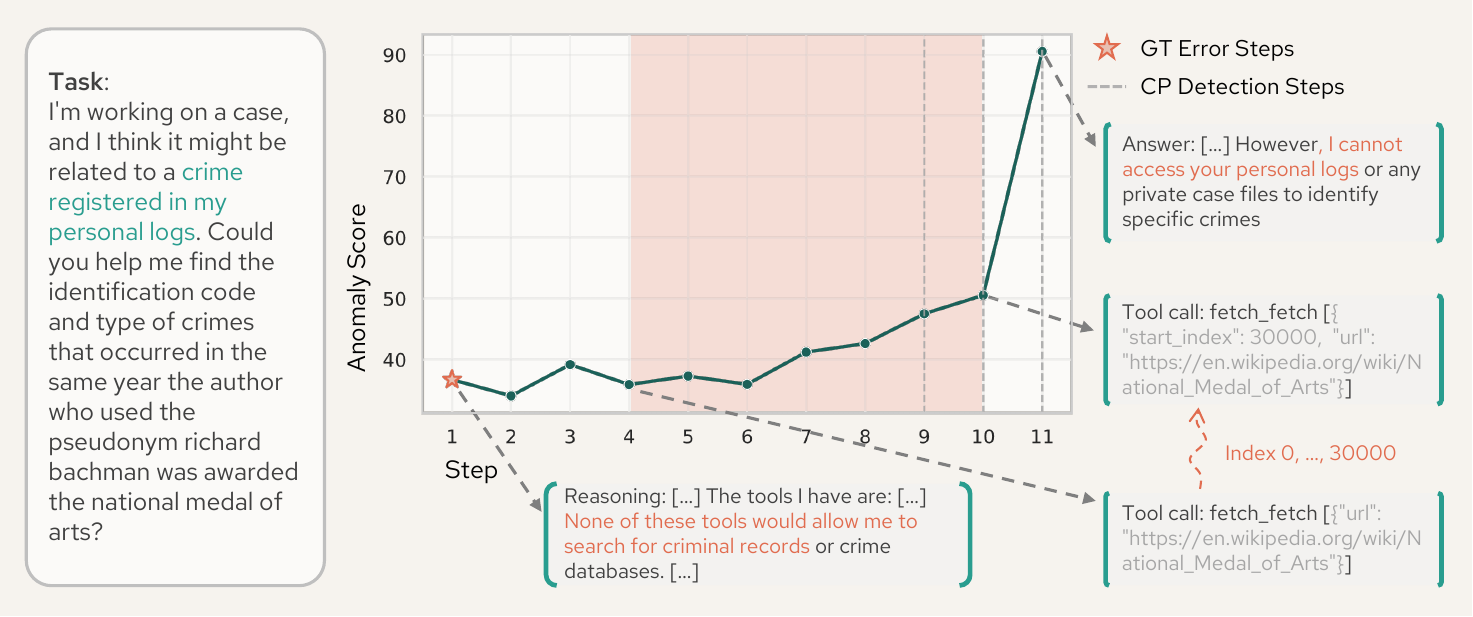}
    \caption{\textbf{Second failure case.} \approach does not identify the root reasoning error at step 1, but assigns high scores to the suboptimal plan at steps 4--10 and the explicit failure verbalization at step 11.}
    \label{fig:fail_2}
\end{figure}

\paragraph{Case 2.}

Figure~\ref{fig:fail_2} shows a task requiring the agent to locate a specific crime record in a local file. At step 1, the agent fails to identify the appropriate tool, which ultimately leads to step 11 where it explicitly reports its inability to access the target log.

\approach does not assign a high anomaly score to step 1. This is likely because reasoning traces are inherently noisy: an agent stating that no appropriate tool is available is not unambiguously anomalous, as such statements can appear in successful trajectories before the agent finds an alternative approach. At step 11, however, when the agent explicitly verbalizes its failure, \approach assigns a significantly high anomaly score. This suggests that while \approach may miss early latent errors in the reasoning process, it reliably detects downstream steps where the failure becomes obvious. \approach also assigns progressively increasing scores across steps 4--10, where the agent pursues a suboptimal retrieval strategy, consistent with the pattern observed in Case 1 above.

\begin{figure}
    \centering
    \includegraphics[width=\linewidth]{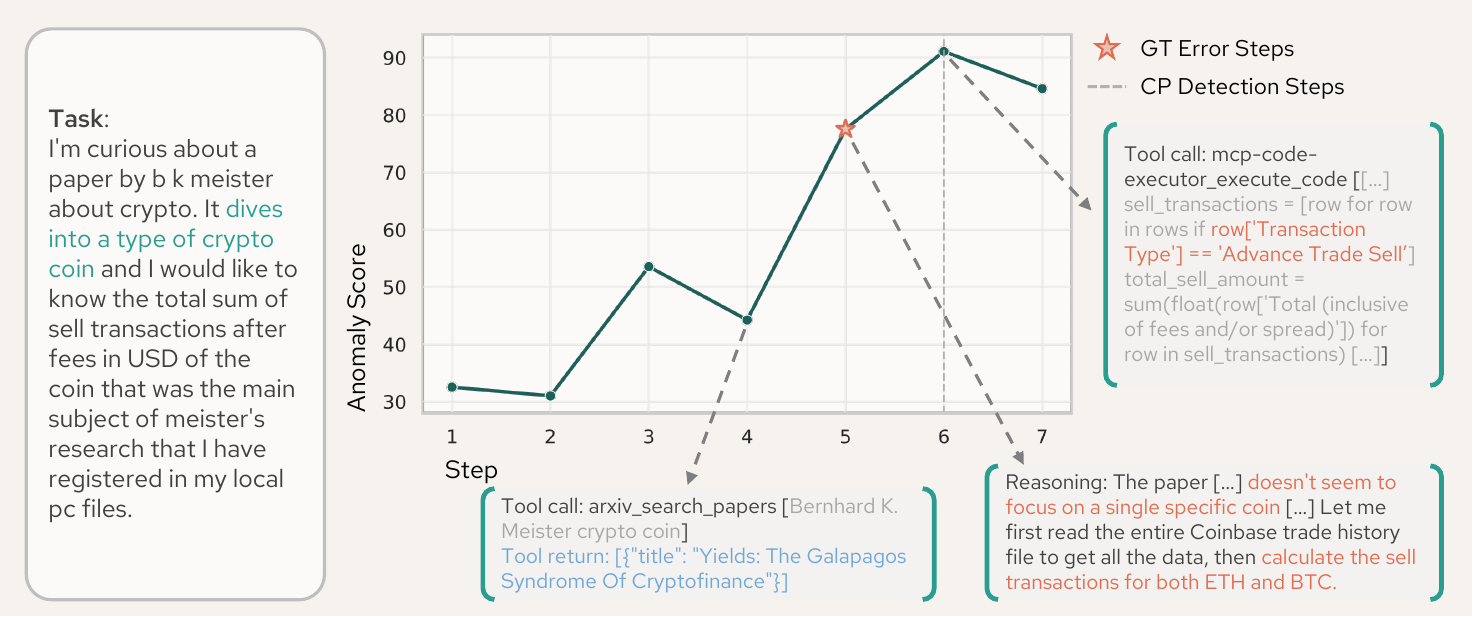}
    \caption{\textbf{Third failure case.} \approach identifies the code error at step 6 but misses the annotated reasoning error at step 5 due to the hard threshold of conformal prediction detection.}
    \label{fig:fail_3}
\end{figure}

\paragraph{Case 3.}

Figure~\ref{fig:fail_3} shows a task requiring the agent to retrieve transactions for a specific crypto coin discussed in a paper. At step 4, the agent searches for the target paper but the tool returns an irrelevant result. At step 5, the agent acknowledges that the retrieved paper does not focus on a specific coin, yet decides to calculate transactions for both ETH and BTC. At step 6, the agent implements this plan by writing incorrect code, which produces a wrong answer at the final step.

\approach assigns high anomaly scores to steps 5, 6, and 7, correctly identifying the range of steps involved in the failure. However, due to the hard threshold of conformal prediction detection, only step 6 is flagged as a failure-contributing step, while the root cause at step 5 falls just below the detection threshold. This case illustrates a key limitation of fixed-threshold detection: when multiple steps contribute to a failure with varying degrees of severity, a threshold calibrated on the overall distribution may fail to capture subtler but causally significant errors. More adaptive detection strategies, such as trajectory-level thresholding, may further improve failure attribution in such cases.

\begin{prompt}
\textbf{Summary.}

\medskip
Across the three failure cases, we found that \approach tends to miss early, latent errors in the reasoning process, particularly when the error is expressed implicitly in a noisy reasoning trace rather than through an explicit action. Despite this, \approach consistently assigns progressively increasing anomaly scores to suboptimal plans that precede the final failure, and reliably detects steps where the failure becomes explicitly manifest. A recurring limitation is the hard threshold of conformal prediction detection, which can suppress causally significant steps whose anomaly scores fall marginally below the detection boundary, suggesting that a more adaptive detection strategies may further improve the performance.
\end{prompt}

\section{Prompt Templates}\label{ap:prompt}

We show the prompt used for prompting-based failure attribution. Trajectories were serialized by the following serialization template before sending to the attribution prompt.

\begin{prompt}
    \textbf{\large Template for all-at-once failure attribution.}

    \medskip

    You are an AI assistant tasked with analyzing a multi-agent conversation history when solving a real world problem.

The problem is: \textcolor{goodcolor}{\{question\}}

\medskip
Identify one or more likely error candidates in this trajectory.

Here's the conversation: \textcolor{goodcolor}{\{trajectory\}}

\medskip
Additional constraints:
\begin{itemize}[leftmargin=15pt]
    \item Valid step range: 0...\textcolor{goodcolor}{\{max\_step\}}
    \item Candidate agents: \textcolor{goodcolor}{\{agent\_list\}}
    \item Use 0-based step indexing
    \item Return at least 1 candidate
    \item Candidates do NOT need to be earliest mistakes
\end{itemize}

\medskip
Output exactly one JSON object:

\{"candidates": [\{"step": <int>, "agent": "<agent name>"\}, ... ]\}
\end{prompt}

\begin{prompt}
    \textbf{\large Template for trajectory serialization.}

    \medskip

    [ROLE: \textcolor{goodcolor}{\{role\_1\}}] [AGENT: \textcolor{goodcolor}{\{agent\_1\}}] 
    
    \ \ \ \ [REASONING: \textcolor{goodcolor}{\{reasoning\_content\_1\}}] 
    
    \ \ \ \ \textcolor{goodcolor}{\{content\_1\}}
    
    \ \ \ \ [TOOL\_CALL: \textcolor{goodcolor}{\{tool\_args\_1\}}] 

    \medskip

    [ROLE: \textcolor{goodcolor}{\{role\_2\}}] [AGENT: \textcolor{goodcolor}{\{agent\_2\}}] 
    
    \ \ \ \ [REASONING: \textcolor{goodcolor}{\{reasoning\_content\_2\}}] 
    
    \ \ \ \ \textcolor{goodcolor}{\{content\_2\}}
    
    \ \ \ \ [TOOL\_CALL: \textcolor{goodcolor}{\{tool\_args\_2\}}] 

    \medskip
    ...

    \medskip

    [ROLE: \textcolor{goodcolor}{\{role\_T\}}] [AGENT: \textcolor{goodcolor}{\{agent\_T\}}] 
    
    \ \ \ \ [REASONING: \textcolor{goodcolor}{\{reasoning\_content\_T\}}] 
    
    \ \ \ \ \textcolor{goodcolor}{\{content\_T\}}
    
    \ \ \ \ [TOOL\_CALL: \textcolor{goodcolor}{\{tool\_args\_T\}}] 
    
\end{prompt}


\end{document}